\documentclass[preprint,authoryear,3p,times,12pt]{elsarticle}

\biboptions{authoryear,round}

\usepackage{geometry}
\usepackage[utf8]{inputenc}
\usepackage[english]{babel}

\usepackage[T1]{fontenc}
\usepackage{microtype}

\usepackage{amsmath,amssymb,amsthm,mathtools}
\usepackage{bm}
\usepackage{mathrsfs}

\usepackage{graphicx}
\usepackage{booktabs}
\usepackage{array}
\usepackage{tabularx}
\usepackage{xcolor}
\usepackage{tikz}
\usepackage{enumitem}
\usepackage{longtable}
\usepackage{pdflscape}
\usepackage{rotating}
\usepackage{float}
\usepackage{placeins}

\usepackage[ruled,vlined,linesnumbered]{algorithm2e}
\DontPrintSemicolon
\SetKwInput{KwInput}{Input}
\SetKwInput{KwOutput}{Output}
\SetKwInput{KwData}{Data}
\SetKwInput{KwResult}{Result}

\usepackage[
    colorlinks=true,
    linkcolor=blue!60!black,
    citecolor=blue!60!black,
    urlcolor=blue!60!black
]{hyperref}

\newtheorem{definition}{Definition}

\newcommand{\argmin}{\operatorname*{arg\,min}}

\newcommand{\Cost}{\mathsf{Cost}}
\newcommand{\Fit}{\mathsf{Fit}}
\newcommand{\Obs}{\mathsf{Obs}}
\newcommand{\Glue}{\mathsf{Glue}}

\newcommand{\C}{\mathcal{C}}

\newcommand{\K}{\mathcal{K}}

\newcommand{\PhiSig}{\Phi}

\newcolumntype{Y}{>{\raggedright\arraybackslash}X}
\newcolumntype{L}[1]{>{\raggedright\arraybackslash}p{#1}}

\usetikzlibrary{arrows.meta,positioning,fit,shapes.geometric,calc}

\makeatletter
\def\ps@pprintTitle{%
  \let\@oddhead\@empty
  \let\@evenhead\@empty
  \def\@oddfoot{\reset@font\hfil\thepage\hfil}%
  \let\@evenfoot\@oddfoot}
\makeatother

\begin{document}

\begin{frontmatter}

\title{Sheaf-Theoretic Transport and Obstruction for Detecting Scientific Theory Shift in AI Agents}

\author{David N. Olivieri}
\ead{olivieri@uvigo.gal}

\author{Roque J. Hern\'andez}
\ead{rhernandez@casagrande.edu.ec}
\address{Universidade de Vigo, Department of Computer Science (LSI), Spain}

\begin{abstract}
Scientific theory shift in AI agents requires more than fitting equations to data. An artificial scientific agent must detect whether an existing representational framework remains transportable into a new regime, or whether its language has become locally-to-globally obstructed and must be extended. This paper develops a finite sheaf-theoretic framework for detecting theory-shift candidates through transport and obstruction. Contexts are organized as a local-to-global structure in which source, overlap, target, and validation charts are fitted, restricted, and tested for gluing. Obstruction measures failure of coherence through residual fit, overlap incompatibility, constraint violation, limiting-relation failure, and representational cost. We evaluate the framework on a controlled transition-card benchmark designed to separate deformation within a source language from extension of that language. The main result is direct obstruction ranking: the intended deformation or extension is usually the lowest-obstruction candidate, and transition type is separated in the benchmark. A constellation kernel over the same signatures is included only as a secondary representational-similarity probe. The aim is not to reconstruct historical paradigm shifts or solve open-ended autonomous theory invention, but to isolate a finite diagnostic subproblem for AI agents: detecting when representational transport fails and extension becomes the coherent next move.
\end{abstract}
 

\end{frontmatter}

\section{Introduction}
\label{sec:introduction}

Scientific cognition depends on the reuse of representations across changing
contexts. A representation developed in one regime may remain valid in another
as an approximation, a limiting case, or a deformed version of the original
model. Classical mechanics survives as a low-velocity limit; ideal-gas
reasoning survives in dilute regimes; small-angle pendulum dynamics survives as
a local chart of a nonlinear system. Scientific understanding therefore
requires more than fitting observations. It requires deciding when an existing
representation can be transported and when the representational language itself
must be enlarged.

Current AI-for-science systems increasingly automate pieces of the scientific
process: equation discovery, program search, hypothesis generation, tool use,
literature search, experiment planning, coding, analysis, and paper-like
reporting. Early computational scientific discovery treated law discovery and
hypothesis search as explicit computational problems
\citep{langley1987scientific}. Modern equation-discovery systems extend this
agenda through symbolic regression, sparse identification of dynamics, and
structured symbolic representations
\citep{schmidt2009distilling,brunton2016sindy,udrescu2020ai_feynman,cranmer2020symbolic}.
Recent benchmark work has further clarified the limits of formula-recovery
tasks and the need to evaluate symbolic discovery systems beyond simple
curve-fitting settings \citep{matsubara2022rethinking_sr_benchmarks}.

A second line of work moves from formula recovery toward interactive and
agentic scientific systems. Program-search systems such as FunSearch show how
language models can participate in mathematically structured exploration
\citep{romera_paredes2024funsearch}. Interactive environments such as
ScienceWorld and DiscoveryWorld evaluate whether agents can plan, experiment,
and infer in simplified scientific worlds
\citep{wang2022scienceworld,jansen2024discoveryworld}. More recent
science-agent benchmarks and autonomous research workflows test data-driven
scientific tasks, workflow execution, and paper-like research loops
\citep{majumder2025discoverybench,chen2025scienceagentbench,lu2024ai_scientist}.
These capabilities sharpen rather than remove a deeper question: whether an AI
agent can recognize when existing representational resources are no longer
adequate, and when genuine scientific progress requires a shift in the theory
language rather than further search within the space of learned patterns.

If an artificial scientific agent is to participate in genuine theory change,
including discovery-like transitions comparable in kind to
Ptolemaic-to-Copernican astronomy, Newtonian-to-Einsteinian mechanics, or
classical-to-quantum transitions, it must diagnose when failure is not merely
poor parameterization or insufficient data, but failure of the representational
language being transported. The question is not whether a system can find a
better-fitting formula inside a supplied search space. The question is whether
it can detect the boundary between deformation inside an existing conceptual
manifold and extension of that manifold. This is the diagnostic boundary
studied here. The paper formulates this boundary in sheaf-theoretic terms: a
source theory, a target regime, and their overlap are treated as local contexts,
and the test is whether the corresponding charts restrict compatibly, glue
across the overlap, preserve limits and constraints, or instead exhibit an
obstruction that motivates extension.

In particular, this paper develops a computational diagnostic for theory
transport and extension. We call the structured object being transported a
\emph{representational constellation}. A constellation includes observables,
law schemas, constraints, measurement roles, limiting regimes, theoretical
posits, and admissible transformations. Galilean velocity addition, for example,
is not only the equation \(w=u+v\); it also carries commitments about absolute
time, unrestricted velocity composition, and the absence of an invariant speed.
Lorentzian velocity composition changes the equation, but also the constraints,
transformations, and limiting relations that define admissible motion.  
The broader motivation is to develop a formal account of scientific
theory shift in the spirit of local-to-global reasoning, with richer
Grothendieck-style and topos-theoretic constructions left for future work. The present paper takes a
smaller step: it asks how an artificial agent can detect, in a finite setting,
when the available local charts no longer glue.

In this paper, a \emph{scientific theory shift} is a transition in which
failure cannot be resolved only by parameter adjustment or bounded deformation
inside the original representational language. A theory shift occurs when
coherence across source, overlap, target, and validation contexts requires a
change in the representational constellation itself: a new primitive,
constraint, law schema, transformation rule, or limiting relation. The central
distinction is therefore between \emph{transport} and \emph{extension}.
Transport preserves the representational language and modifies the
representation within that language. Extension changes the language by adding
one of these representational resources. In future work, richer categorical
machinery may provide a way to compare or relate obstructed theoretical
contexts through pullback-like constructions; here, however, the objective is
deliberately limited to the finite diagnostic problem of detecting when
transport fails and extension is required.

Open problems such as dark matter or dark energy illustrate the scale of the
challenge. The issue in such cases is not simply to fit one curve, but to assess
whether an extended constellation of laws, constraints, scales, measurements,
and auxiliary assumptions remains coherent across domains. The present
benchmark is far smaller and does not attempt to model such cases. It isolates
a finite version of the same diagnostic question for AI agents: when does
representational transport fail?

Although the definitions introduced here are operational and computational,
they are consistent with several conceptual traditions. In Popperian terms,
failures matter, but prediction failure alone does not determine whether a
problem is local, parametric, or representational \citep{popper1959logic}.
Kuhnian and cognitive accounts emphasize that major scientific shifts involve
changes in representational resources, not only better numerical fits
\citep{kuhn,thagard2012,nersessian2008creating}. Work on theory building also
stresses that scientific construction uses heuristics, case studies, cognitive
processes, and defensible but non-guaranteed principles
\citep{danks2018building_theories}. Our framework also connects to cognitive-systems accounts in which
representational change and knowledge generation are modeled as computational
mechanisms \citep{sun2009theoretical,lieto2019beyond}.

The formalism below makes this diagnostic boundary finite and computable. In
standard sheaf theory, local data are assigned to contexts, restriction maps
compare descriptions across refinements, and compatible local sections can glue
into a coherent global section \citep{maclane_moerdijk,johnstone}. Here this
structure is instantiated operationally: contexts are source, overlap, target,
and validation regimes; local charts are fitted representational
constellations; restriction evaluates charts on shared overlap observations;
gluing measures compatibility; and obstruction measures failure to fit, glue,
preserve limits, or satisfy constraints. The construction is finite and
local-to-global. It is motivated by sheaf-theoretic ideas, but it is not a
computation in full topos semantics.

The hypothesis adopted here is that discovery-like revision begins where
representational transport fails. If a source constellation can be deformed
within its original language so that it fits the target, agrees on overlaps,
satisfies constraints, and preserves the source limit, the transition is a case
of transport. If no bounded deformation removes the local-to-global
obstruction, the system must search for a minimal extension of the
representational language. Such an extension is not merely extra flexibility;
it is a justified change in the representation that makes the source, overlap,
and target regimes coherent again.

We test this hypothesis with a controlled benchmark built from
\emph{transition cards}. A transition card is a finite, structured record of
one proposed theory shift: it specifies a source constellation, the regimes in
which that constellation is tested, the observations available in source,
overlap, target, and validation contexts, and a finite set of candidate
representational moves. In this sense, a card is a small computational object
for asking whether a source representation can be transported into a new regime
or whether the representation itself must be extended. The benchmark contains
physics-inspired transition families of two kinds. In deformation-sufficient
cards, the correct move remains inside the original representational language.
In extension-required cards, the correct move introduces a new primitive,
constraint, law schema, transformation rule, or limiting relation. The
experiment evaluates whether obstruction signatures rank the intended
representational move and whether gluing information contributes to
distinguishing transport from representational strain.

This paper takes an initial computational step toward a broader program on
genuine scientific discovery in AI agents. It does not attempt to solve
open-ended autonomous theory invention or to reconstruct historical paradigm
shifts. Instead, it isolates a necessary diagnostic subproblem: when a familiar
theory is moved outside its native regime, can an artificial scientific agent
detect whether deformation is sufficient, or whether extension of the
representational language is required?

\subsection{Contributions}
\label{subsec:contributions}

This paper makes four contributions. First, it casts scientific theory shift
as a finite diagnostic problem for AI agents: detecting when transport inside a
source representational language is insufficient and extension is required.
Second, it introduces representational constellations as structured local charts
for scientific models. Third, it formalizes transport, restriction, gluing,
obstruction, and minimal extension in a finite sheaf-theoretic setting. Fourth,
it evaluates the resulting obstruction signatures, together with a secondary
constellation-kernel probe, on controlled transition families that separate
deformation from extension.

\section{Sheaf-Theoretic Background for Local-to-Global Structure}
\label{sec:sheaf_background}

The mathematical background of the paper is sheaf theory. The relevant idea is local-to-global organization: data, descriptions, or constraints are assigned locally over a base of contexts, and a global description exists only when the local descriptions are compatible on their overlaps. This is the classical role of sheaves in geometry and logic \citep{maclane_moerdijk,johnstone}. Applied sheaf theory uses the same local-to-global structure for data fusion and networked consistency problems \citep{curry2014,robinson2017}, while finite and cellular sheaf methods make compatibility, inconsistency, and obstruction computationally tractable \citep{hansen_ghrist2019,ayzenberg2025}.

This local-to-global viewpoint is natural for scientific representations. A theory is often valid first as a local chart over a domain of applicability: a small-angle pendulum model applies over small angular amplitudes, Newtonian kinetic energy in a low-velocity regime, and Rayleigh--Jeans reasoning in a long-wavelength or low-frequency regime \citep{feynman2011lectures}. In each case, the representation carries not only a predictive formula but also a domain of validity, a set of constraints, and a relation to neighboring regimes. The question is therefore not only whether a model fits a set of observations, but whether local descriptions can be restricted, compared, and glued into a coherent larger description.

\subsection{Contexts, refinements, and covers}

Let \(\C\) be a finite category of contexts. An object \(U\in\C\) represents a regime in which observations, constraints, and representational assumptions are specified. In the present paper, a context is a regime of use: source, overlap, target, validation, or more generally any domain in which a description is meant to apply. For a scientific agent, a context is therefore not just a data subset; it is a regime in which a model is being asked to make sense under particular assumptions. A context may encode a physical regime, an approximation domain, a measurement protocol, a data-quality restriction, or a modeling assumption.

A morphism
\[
    V \longrightarrow U
\]
is interpreted as a refinement or restriction of context. It means that \(V\) is a more specific or more demanding view of \(U\): a narrower domain, a stricter measurement setting, a higher-resolution regime, or a target regime in which additional constraints become active. In scientific terms, this is the operation of asking how a description valid in one regime behaves when read in a more specific or overlapping regime. This is the same formal role played by restriction maps in sheaf theory and by change-of-context maps in sheaf semantics \citep{maclane_moerdijk,johnstone,caramello}.

A cover of a context \(U\) is a family of local contexts
\[
    \{U_i \longrightarrow U\}_{i\in I}
\]
that jointly probe \(U\). Covers express the idea that a broader regime can be studied through compatible local regimes. For scientific discovery, this means that a theory is not tested in one undifferentiated domain, but across partial regimes whose compatibility must be checked. In the finite experiments below, source, overlap, and target regimes form the operational cover used to test whether a candidate representation can be treated as one coherent description across regimes. The overlap context is central because it is where independently fitted local descriptions are restricted and compared. 

As a running example, consider transporting Galilean velocity composition into a higher-velocity regime. The source context \(U_s\) contains low-velocity observations where the additive law \(w=u+v\) is adequate. The target context \(U_t\) contains higher subluminal velocities where invariant-speed constraints become relevant. The overlap context \(U_o\) contains intermediate velocities where source-fitted and target-fitted descriptions can both be evaluated. In the finite site used here, \(U_o\) is treated as a common refinement of the source and target regimes, with maps \(U_o\to U_s\) and \(U_o\to U_t\). These maps express that the overlap is the common regime on which the two local descriptions must be restricted and compared. The question is whether these restricted descriptions agree sufficiently to be treated as one transported representation, or whether the mismatch indicates obstruction.

\subsection{Presheaves as context-dependent descriptions}

Intuitively, a presheaf organizes descriptions that depend on context. To connect this intuition with standard notation, let \(X\) denote a space of contexts or regimes, and let \(\mathcal{O}(X)\) be a family of admissible contexts. A presheaf of representations is written
\[
    \mathcal{F}:\mathcal{O}(X)^{\mathrm{op}}\to \mathbf{Set},
    \qquad
    U\mapsto \mathcal{F}(U),
\]
where \(\mathcal{F}(U)\) is the set of admissible descriptions, local laws, or representational constellations on context \(U\). For an inclusion or refinement \(V\subseteq U\), the restriction map is
\[
    \rho^U_V:\mathcal{F}(U)\to \mathcal{F}(V).
\]
This is the standard contravariant notation for a presheaf \citep{maclane_moerdijk,johnstone}. In scientific terms, the restriction map answers the question: if a description is valid in one regime, what does that same description imply when it is read, evaluated, or compared in a more
specific or overlapping regime?

Here, \(X\) is not an arbitrary topological space. The operational site is finite: source, overlap, target, and validation regimes play the role of contexts; candidate constellations play the role of local sections; and restriction is implemented by evaluating fitted charts on shared overlap observations. A presheaf is therefore the bookkeeping structure that records which representational constellations are admissible in each regime and how those constellations are compared across regimes. Because empirical compatibility is approximate, the exact sheaf condition is replaced below by a quantitative obstruction.

Applied sheaf theory uses this structure to represent distributed measurements, local constraints, and compatibility across networks or graphs \citep{robinson2017,curry2014,hansen_ghrist2019}. In knowledge representation, sheaf-theoretic formulations describe embeddings or assignments as approximate global sections satisfying local schema constraints \citep{gebhart2023}. In graph learning, cellular sheaves generalize graph-based diffusion by assigning structured data and restriction maps to nodes and edges rather than treating the graph as a homogeneous carrier of scalar features \citep{bodnar2022}. The common theme is that local assignments matter not only individually, but also through the way they restrict and agree across shared structure.

The key point for scientific representation is that models are not compared only by global prediction error. They must also behave correctly under restriction. A representation that fits a target regime but destroys the source limit has not transported the source theory correctly. A representation that fits two local regimes but gives incompatible consequences on their overlap has not produced a coherent global chart.

Continuing the Galilean-to-Lorentz example, \(\mathcal{F}(U_s)\) contains descriptions admissible in the low-velocity source regime, including the Galilean additive law. The target set \(\mathcal{F}(U_t)\) contains descriptions admissible in the higher-velocity regime, where invariant-speed constraints may become active. The overlap set \(\mathcal{F}(U_o)\) contains descriptions evaluated in the intermediate regime. Restriction maps such as \(\rho^{U_s}_{U_o}\) and \(\rho^{U_t}_{U_o}\) express how source-fitted and
target-fitted descriptions are read on the common overlap. The obstruction test will ask whether these restricted descriptions agree as one transported
constellation, or whether their mismatch indicates that the original presheaf of admissible descriptions must be extended.

\subsection{The sheaf condition: agreement and gluing}

A sheaf is a presheaf with an additional local-to-global property. If several
local descriptions agree when restricted to their overlaps, then they should be
understood as parts of one coherent global description. This is the sense in
which gluing is used here. In scientific terms, the point is simple: it is not
enough for a model to work separately in several nearby regimes; those local
uses must also agree where the regimes meet.

Suppose a context \(U\) is covered by local contexts
\[
    \{U_i \to U\}_{i\in I}.
\]
A local description is a section \(s_i\in\mathcal{F}(U_i)\). In the present
setting, one can think of \(s_i\) as a candidate representational
constellation---a law schema together with its constraints and limiting
assumptions---as used in regime \(U_i\). A family \(\{s_i\}_{i\in I}\) is
compatible, or a matching family, when its restrictions agree on pairwise
overlaps:
\[
    \rho^{U_i}_{U_i\cap U_j}(s_i)
    =
    \rho^{U_j}_{U_i\cap U_j}(s_j)
    \qquad \text{for all } i,j.
\]
Compatibility means that the local descriptions do not contradict one another
on their shared domains. They may have been fitted or formulated locally, but
once restricted to the common regime, they make the same claims there.

A sheaf is a presheaf in which every compatible family of local sections glues
to a unique section \(s\in\mathcal{F}(U)\) satisfying
\[
    \rho^U_{U_i}(s)=s_i
    \qquad \text{for all } i\in I.
\]
This is the standard locality-and-gluing condition for sheaves
\citep{maclane_moerdijk,johnstone}. Intuitively, if the local pieces already
agree wherever they overlap, then there is one coherent global description of
which they are the local parts.

This is the formal local-to-global principle used throughout the paper. Local
adequacy is not enough. A candidate representation must also glue. In
scientific terms, a model should fit the source context, fit the target
context, preserve the correct limiting relation, and give compatible
consequences on the overlap. A candidate representation can fit source and
target observations separately, but if its source-fitted and target-fitted
charts disagree on the overlap, then the local pieces do not glue. Similar
local-to-global consistency ideas appear in sheaf models of contextuality
\citep{abramsky2011}, distributed task solvability \citep{felber2025}, and
sensor integration \citep{robinson2017}.

\begin{figure}[t]
    \centering
    \includegraphics[width=\textwidth]{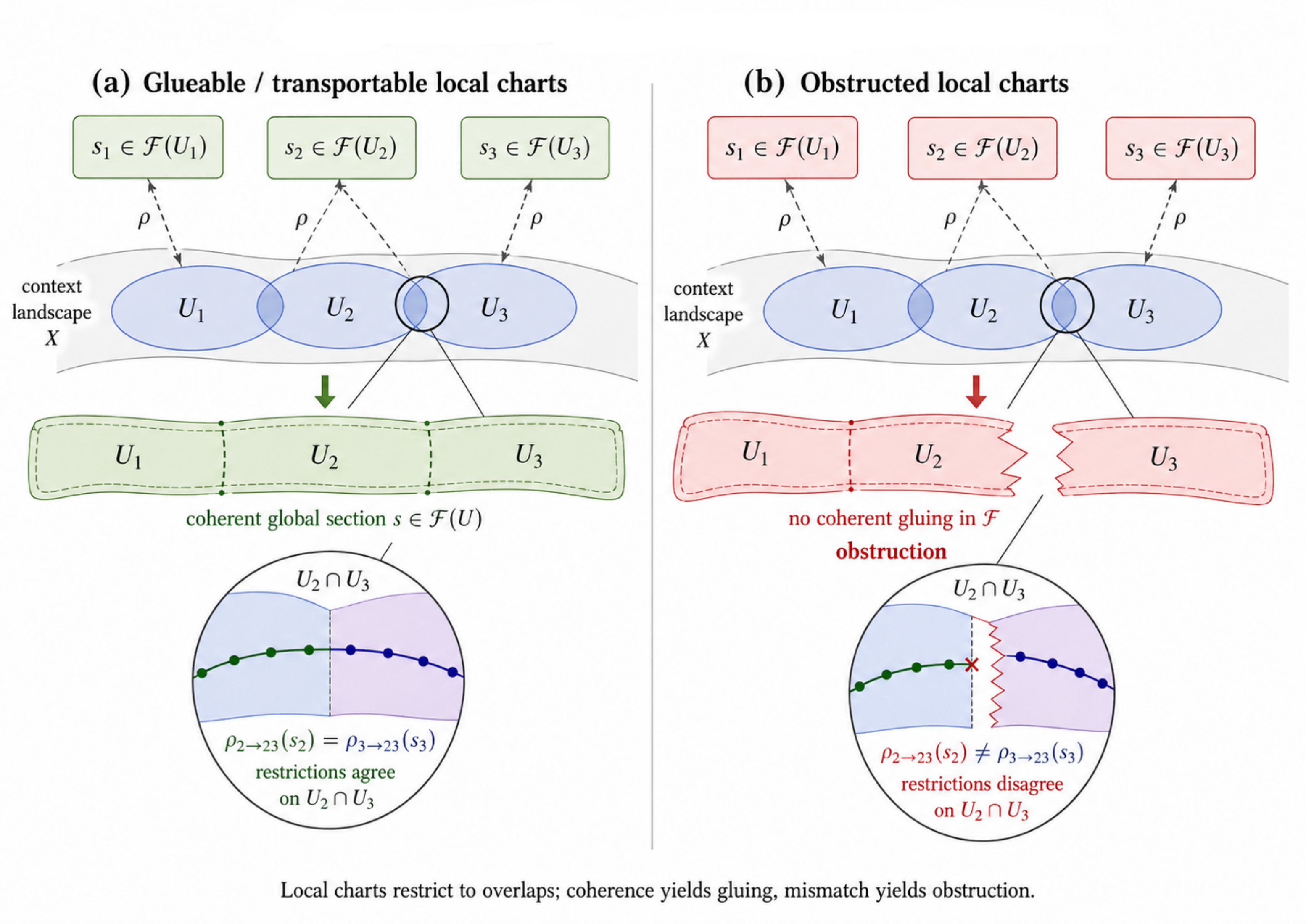}
    \caption{Geometric intuition for restriction, gluing, and obstruction. The context landscape is covered by overlapping local regions \(U_1,U_2,U_3\), each equipped with a local section \(s_i \in \mathcal{F}(U_i)\). In panel (a), the restrictions of the local sections agree on overlaps, so the local descriptions can be glued into a coherent global section \(s \in \mathcal{F}(U)\). In panel (b), the restrictions disagree on an overlap, producing a finite local-to-global obstruction within the present representational family. The drawing is schematic: the paper operationalizes this local-to-global idea with finite source, overlap, target, and validation contexts rather than a full topological sheaf.}
    \label{fig:sheaf_gluing_intuition}
\end{figure}

Figure~\ref{fig:sheaf_gluing_intuition} gives the geometric intuition behind
the finite construction used below. In panel (a), the local charts
\(s_1,s_2,s_3\) restrict compatibly on the overlaps, so they can be interpreted
as parts of one coherent global section. In panel (b), the local charts still
exist and may each be locally meaningful, but the restricted descriptions fail
to agree on the shared overlap, so no coherent gluing is available inside the
same representational family. The bottom insets visualize this difference:
agreement on the overlap yields a smooth continuation, whereas disagreement
produces a visible mismatch.

As a running scientific example, consider again the attempt to transport
Galilean velocity composition into a higher-velocity regime. Let \(U_1\) denote
a low-velocity source regime, \(U_3\) a higher-velocity target regime, and
\(U_2\) an intermediate overlap regime. A source-based local chart \(s_1\) may
encode the additive law \(w=u+v\), while a target-based chart \(s_3\) may
encode a representation adapted to the higher-velocity regime. The overlap
chart \(s_2\) represents the intermediate domain in which both sides can be
compared. If the restrictions of these descriptions agree on the shared regime,
then they behave like panel (a) of Figure~\ref{fig:sheaf_gluing_intuition}: the
theory has been transported coherently. If the restrictions disagree, as in
panel (b), then the mismatch indicates that the source representation does not
extend coherently across regimes. In the finite setting of this paper, that
failure of gluing is what motivates the introduction of obstruction.

\subsection{Obstruction as failed gluing}

In the exact sheaf condition, compatibility is categorical: local sections
either agree on overlaps or they do not. In empirical settings, compatibility
is approximate and quantitative. Data are noisy, models are approximate, and
agreement has to be measured rather than asserted. We therefore use a finite
obstruction functional. Given a candidate constellation \(K\), local contexts
\(U_i\), and overlaps \(U_i\cap U_j\), we measure how far the locally fitted
descriptions are from gluing:
\[
    \Glue(K)
    =
    \sum_{i<j}
    d\!\left(
        \rho^{U_i}_{U_i\cap U_j}(\widehat K_i),
        \rho^{U_j}_{U_i\cap U_j}(\widehat K_j)
    \right),
\]
where \(\widehat K_i\) denotes the candidate fitted in context \(U_i\), and
\(d\) is a discrepancy between restricted predictions or constraint profiles.
In scientific terms, \(\Glue(K)\) asks whether independently fitted local
versions of the same proposed representation make compatible claims where
their regimes overlap.

This use of obstruction follows the applied sheaf-theoretic view that
incompatibility of local data is itself informative. Cellular sheaf Laplacians
and sheaf cohomology provide computational ways to quantify consistency,
disagreement, and obstruction in finite settings
\citep{curry2014,hansen_ghrist2019,ayzenberg2025}. The benchmark uses the same
principle in a simpler finite form: obstruction is a measured failure of local
descriptions to become one coherent representation. In the theory-shift setting,
such failure is not treated merely as residual error. It is evidence that the
current representational family may not transport across regimes.

The full obstruction also includes local residuals, constraint violations,
limit failures, and representational cost. Thus obstruction is not simply
prediction error. It measures failure of local-to-global coherence:
\[
    \Obs(K)
    =
    w_{\mathrm{res}} R_{\mathrm{res}}(K)
    +
    w_{\mathrm{glue}} G_{\mathrm{glue}}(K)
    +
    w_{\mathrm{con}} C_{\mathrm{viol}}(K)
    +
    w_{\mathrm{lim}} P_{\mathrm{limit}}(K)
    +
    \lambda\,\Cost(K).
\]
Here \(R_{\mathrm{res}}(K)\) aggregates local residuals across source, overlap,
target, and validation contexts; \(G_{\mathrm{glue}}(K)\) measures disagreement
between restricted local charts; \(C_{\mathrm{viol}}(K)\) penalizes violations
of structural constraints; \(P_{\mathrm{limit}}(K)\) penalizes failure to
recover the source representation as an appropriate limit; and \(\Cost(K)\)
penalizes unnecessary representational change.

A low-obstruction candidate is one whose local descriptions fit, restrict, and
glue while preserving the relevant constraints and limiting relations. A
high-obstruction candidate may fit some local data but fails as a coherent
representation across contexts. In the Galilean-to-Lorentz example, this means
that a candidate should not only fit low- and higher-velocity observations
separately; its source- and target-fitted charts should also agree on the
intermediate overlap and preserve the relevant low-velocity limit. Failure on
these terms is what the obstruction functional records.

\subsection{Transport and extension}

The distinction between transport and extension can now be stated in
sheaf-theoretic terms. Transport tries to carry a description from one context
to another while keeping the same representational language. In the notation
above, transport asks whether a source description
\(s\in\mathcal{F}(U_s)\) can be restricted, deformed, or refitted so that it
remains compatible with the overlap and target contexts. In scientific terms,
transport is the case in which the original language still has enough expressive
resources to make sense of the new regime.

Extension changes the presheaf itself. A new relation, constraint, primitive,
law schema, or limiting relation enlarges the set of admissible descriptions:
\[
    \mathcal{F} \quad \leadsto \quad \mathcal{F}^{+}.
\]
The extension is justified when obstruction remains high inside
\(\mathcal{F}\), but drops after passing to \(\mathcal{F}^{+}\). In other
words, the system does not add expressive capacity merely to improve fit; it
adds a specific representational resource because that resource restores
coherence across source, overlap, and target regimes.

In the Galilean-to-Lorentz example, transport would mean that the Galilean
representational language can be deformed or refitted so that its low-velocity
law, overlap behavior, and higher-velocity predictions remain mutually
compatible. Extension is required when this cannot be done inside the original
language. Adding an invariant-speed constraint and Lorentzian velocity
composition changes the admissible descriptions from \(\mathcal{F}\) to an
extended family \(\mathcal{F}^{+}\), within which the source limit and
higher-velocity regime can be made coherent.

Thus, the sheaf-theoretic structure gives the paper its central criterion. A
representational shift is a deformation when coherence can be restored inside
the original presheaf of descriptions. It is an extension when coherence
requires enlarging that presheaf. This is why the benchmark below separates
deformation-sufficient transition families from extension-required transition
families.

\subsection{Finite sheaf model used in this paper}

The experiments instantiate the preceding definitions on a finite site. Each
transition card supplies four contexts,
\[
    U_s,\quad U_o,\quad U_t,\quad U_v,
\]
corresponding to source, overlap, target, and validation regimes. A candidate
move \(\Delta_j\) produces a candidate constellation \(K_j=K_0+\Delta_j\).
The associated presheaf \(\mathcal{F}\) specifies which descriptions are
admissible in each context, and the fitted local charts
\[
    \widehat K_{j,s}\in\mathcal{F}(U_s),
    \qquad
    \widehat K_{j,t}\in\mathcal{F}(U_t)
\]
are restricted to the overlap by evaluating them on \(U_o\). Gluing is then the
measured disagreement between
\[
    \rho^{U_s}_{U_o}(\widehat K_{j,s})
    \quad\text{and}\quad
    \rho^{U_t}_{U_o}(\widehat K_{j,t}),
\]
while obstruction combines this disagreement with residual fit, constraint
penalties, limit penalties, and representational cost.

In the Galilean-to-Lorentz running example, \(U_s\) contains low-velocity data,
\(U_o\) contains intermediate velocities where both source- and target-fitted
charts can be tested, \(U_t\) contains higher subluminal velocities, and \(U_v\)
provides held-out validation. A fixed Galilean candidate may fit \(U_s\), but
its restrictions can disagree with the target-fitted chart on \(U_o\) and fail
the invariant-speed constraints in \(U_t\). A Lorentzian extension changes the
admissible constellation family so that the overlap restrictions agree, the
low-velocity limit is preserved, and the higher-velocity constraints are
satisfied. This finite computation is the operational form of transport,
gluing, and obstruction used in the benchmark.

The construction is finite and computational, but it follows the same
local-to-global logic used in applied sheaf work on data fusion and distributed
consistency, where local assignments are compared through restriction maps and
global coherence becomes a computable property
\citep{robinson2017,felber2025}. Related work in knowledge representation and
graph learning likewise treats local assignments, schemas, or cellular sheaf
data as objects whose compatibility can be computed
\citep{gebhart2023,bodnar2022,ayzenberg2025}. The benchmark below uses this
logic as a diagnostic for scientific theory shift rather than as a full
topos-theoretic semantics.

\section{Representational Constellations}
\label{sec:constellations}

Scientific representations are not exhausted by their equations. A scientific model also carries a structured set of commitments: which quantities are observable, which entities are theoretical posits, which transformations are admissible, which constraints must be preserved, which measurements define the relevant variables, and which limiting regimes connect the model to neighboring descriptions. This view is consistent with work on models as mediators in scientific practice \citep{morgan_morrison1999}, model-based reasoning and conceptual change \citep{nersessian2008creating,thagard2012}, and conceptual spaces as structured representational resources
\citep{gardenfors2000conceptual}.

We call this organized structure a \emph{representational constellation}. A constellation is a local chart for a scientific regime: it contains law schemas, observables, measurement roles, constraints, limit relations, transformation rules, and theoretical posits. The term ``constellation'' emphasizes that the object being transported is not a single formula, but a configuration of representational elements that jointly determine what counts as an admissible description.

For example, Galilean velocity addition is not only the equation \(w=u+v\). It belongs to a constellation involving absolute time, unconstrained velocity composition, Galilean transformations between inertial frames, and the absence of an invariant speed. Lorentzian velocity composition changes the formula, but also the constraints, transformations, and limiting relations that organize admissible motion. Likewise, the Rayleigh--Jeans law is not merely a failed high-frequency formula; it belongs to a classical equipartition constellation without a quantization scale. Planck-like radiation introduces a new primitive  and a different admissibility structure. In these cases, conceptual change is not simply parameter replacement, but reorganization of the representational constellation \citep{kuhn,nersessian2008creating,thagard2012}.

\begin{table}[t]
\centering
\small
\caption{Representational constellation shift from Galilean to Lorentzian velocity composition. The shift changes not only the composition law, but also the constraints, transformations, and limiting structure that define admissible motion.}
\label{tab:galilean_lorentz_constellation}
\begin{tabular}{p{0.22\linewidth}p{0.34\linewidth}p{0.34\linewidth}}
\toprule
Component & Galilean constellation & Lorentzian constellation \\
\midrule
Observable quantities
& Relative velocities \(u,v,w\)
& Relative velocities \(u,v,w\) interpreted relative to invariant speed \(c\) \\

Law schema
& Additive composition:
\[
    w = u+v
\]
& Relativistic composition:
\[
    w=\frac{u+v}{1+uv/c^2}
\] \\

Theoretical posit
& Absolute time; no invariant finite speed
& Invariant light speed; frame-independent speed bound \\

Admissible transformations
& Galilean transformations between inertial frames
& Lorentz transformations between inertial frames \\

Structural constraints
& Velocity composition is unbounded
& Composed velocities remain bounded by \(c\) when \(|u|,|v|<c\) \\

Limit relation
& No source-local limiting relation to a broader velocity law
& Recovers Galilean composition when \(|u|/c\ll 1\) and \(|v|/c\ll 1\) \\

Representational role
& Source constellation
& Extension adding invariant-speed structure, Lorentz transformations, and a source-regime limit \\
\bottomrule
\end{tabular}
\end{table}

\subsection{Constituents of a constellation}

A representational constellation \(\K\) is modeled as a typed structure
\[
    \K =
    \bigl(
    \mathcal O,
    \mathcal P,
    \mathcal L,
    \mathcal C_{\mathrm{str}},
    \mathcal M,
    \mathcal R_{\mathrm{lim}},
    \mathcal T
    \bigr),
\]
where \(\mathcal O\) denotes observables, \(\mathcal P\) theoretical posits, \(\mathcal L\) law schemas, \(\mathcal C_{\mathrm{str}}\) structural constraints, \(\mathcal M\) measurement roles, \(\mathcal R_{\mathrm{lim}}\) limit relations, and \(\mathcal T\) admissible transformations. This tuple is a working representation rather than a complete ontology of science: it keeps the features needed to distinguish deformation within a representational language from extension of that language.

Each component has a different diagnostic role. Observables define what can be compared to data. Law schemas specify functional or relational structure. Structural constraints encode admissibility conditions such as conservation, boundedness, positivity, finite-energy behavior, or invariant-speed conditions. Limit relations specify how one description reduces to another in a source regime. Transformation rules specify how quantities may change across frames, scales, or contexts. Measurement roles connect theoretical quantities to the observations that instantiate them.

Separating these components matters because failure can occur in different places. A candidate formula may fit a target regime but violate the source limit. Another may fit source and target observations but fail to agree on the overlap. Another may lower residual error by adding flexible terms while violating the structural constraint that made the original representation meaningful. Treating a model as a constellation makes these failures explicit rather than collapsing them into a single prediction-error score.

\subsection{Constellations as typed graphs}

For computation, a constellation is encoded as a typed graph
\[
    G_{\K}=(V_{\K},E_{\K},\tau_V,\tau_E),
\]
where \(V_{\K}\) contains representational elements, \(E_{\K}\) contains relations among them, and \(\tau_V,\tau_E\) assign node and edge types. The node-type map \(\tau_V\) distinguishes \emph{observables}, \emph{theoretical posits}, \emph{law schemas}, \emph{constraints}, \emph{measurement roles}, \emph{limit relations}, \emph{transformation rules}, and \emph{contexts}. The edge-type map \(\tau_E\) distinguishes relations of \emph{use}, \emph{assumption}, \emph{constraint}, \emph{preservation}, \emph{validity in a context}, \emph{reduction to a limit}, \emph{extension}, \emph{conflict}, \emph{measurement}, \emph{introduction}, and \emph{removal}.

The graph does not replace the mathematical law schema. Rather, if \(\ell\in V_{\K}\) is a node of type \emph{law schema}, the incident typed edges record the commitments surrounding \(\ell\): assumptions \((\ell,\mathrm{assumes},p)\), constraints \((\ell,\mathrm{constrains},c)\), contexts \((\ell,\mathrm{valid\_in},U)\), limiting relations \((\ell,\mathrm{preserves},r)\), and representational changes such as \((\ell,\mathrm{introduces},q)\) or \((\ell,\mathrm{removes},q')\). This is why two candidates with similar residual error may still be different constellations: one may preserve a source limit while another breaks it; one may introduce a new primitive while another merely deforms an old formula.

This representation also makes transition families comparable. The graph is not intended as a generic knowledge graph, but as a typed record of the local scientific chart being tested. It keeps the connection to scientific models as structured representational resources \citep{gardenfors2000conceptual,morgan_morrison1999}, while giving the benchmark a computable object for structural comparison. In the experiments, these graph records support secondary similarity probes through graph-based representations and graph kernels \citep{vishwanathan2010}.

\subsection{Deformation and extension}

A change of constellation can preserve the existing representational language or enlarge it. A \emph{deformation} modifies a law schema, parameterization, or
correction term while keeping the same basic representational resources: 
\[
    \K \leadsto \K_{\theta}.
\]
Here \(\K\) is the source constellation and \(\theta\) denotes an admissible within-language adjustment. For example, if \(\K\) contains the small-angle pendulum law, then \(\K_{\theta}\) may contain a finite-angle correction parameterized by amplitude. If \(\K\) contains the ideal-gas law, then \(\K_{\theta}\) may add virial coefficients. If \(\K\) contains Ohm's law with constant resistance, then \(\K_{\theta}\) may add temperature dependence. In each case, the original variables, constraints, and limiting interpretation remain recognizable, and the source description is preserved as an appropriate limit.

An \emph{extension} changes what counts as an admissible description:
\[
    \K \leadsto \K^{+}.
\]
Here \(\K^{+}\) is not just a reparameterized version of \(\K\); it is an enlarged constellation with a new primitive, constraint, transformation rule, limiting relation, or theoretical posit. If \(\K\) is the Galilean velocity constellation, then \(\K^{+}\) may add invariant-speed structure, Lorentz transformations, and Lorentzian velocity composition. If \(\K\) is the Newtonian kinetic-energy constellation, then \(\K^{+}\) may add a relativistic high-velocity law with the Newtonian expression as a limit. If \(\K\) is the Rayleigh--Jeans constellation, then \(\K^{+}\) may add a quantization scale and a finite-energy constraint. These moves are not merely larger parameter spaces. They change the representational resources available to the model.

This distinction is the one tested in the experiments. For a transition card \(T\), candidate moves generate constellations
\[
    \K_j=\Delta_j(\K).
\]
A card is \emph{deformation-sufficient} when the lowest-obstruction candidate belongs to the within-language family \(\{\K_{\theta}\}\). It is \emph{extension-required} when every admissible deformation remains obstructed and the lowest-obstruction candidate lies in an enlarged family \(\mathcal F^{+}\), represented by some \(\K^{+}\). Thus the benchmark does not ask only which candidate fits the target best; it asks whether low obstruction is achievable inside the source language or only after the language is
extended.

\subsection{Constellations as local charts}

A constellation \(\K\) is indexed by a context \(U\). It is not simply true or false globally; it is valid over a regime, under specified measurement roles, constraints, and limiting assumptions. A fitted constellation in context \(U\) is therefore treated as a local chart
\[
    \widehat{\K}_{U}\in\mathcal F(U),
\]
where \(\mathcal F(U)\) is the set of admissible constellations in that context. The sheaf-theoretic question is whether charts fitted in different contexts can be restricted to their overlaps and glued into a coherent larger chart. 

For a transition card \(T\), formally defined in the next section, the source constellation \(\K_0\) is tested against source, overlap, target, and validation observations. Each candidate move \(\Delta_j\) acts on the source constellation to produce a candidate constellation \(\K_j=\Delta_j(\K_0)\). Fitting \(\K_j\) in the source and target regimes gives local charts
\[
    \widehat{\K}_{j,s}\in\mathcal F(U_s),
    \qquad
    \widehat{\K}_{j,t}\in\mathcal F(U_t).
\]
These charts are compared by their restrictions to the overlap,
\[
    \rho^{U_s}_{U_o}(\widehat{\K}_{j,s})
    \quad\text{and}\quad
    \rho^{U_t}_{U_o}(\widehat{\K}_{j,t}).
\]
In the Galilean-to-Lorentz example, this asks whether the low-velocity source-fitted chart and the higher-velocity target-fitted chart make compatible claims in the intermediate velocity regime. Agreement supports transport; disagreement contributes to obstruction.

This local-chart view connects the representational object \(\K_j\) to the finite local-to-global test used in the experiments. Section~\ref{sec:formalism} turns the comparison above into an obstruction functional: a candidate transports when its fitted charts agree on overlaps while preserving constraints and limits, and it requires extension when low obstruction is attainable only after enlarging the representational constellation.

\section{Transport, Gluing, and Obstruction Formalism}
\label{sec:formalism}

The experiments use a finite site of scientific contexts and a scalar obstruction functional. Contexts form the site, constellations are fitted as local charts, restriction compares those charts on overlaps, gluing measures their compatibility, and obstruction quantifies failure of local-to-global coherence \citep{maclane_moerdijk,johnstone}. This finite construction follows the computational direction of applied sheaf work on data fusion, consistency, and cellular sheaf methods
\citep{curry2014,robinson2017,hansen_ghrist2019}.

\subsection{Finite site of scientific contexts}

Let \(\C\) be a finite category of scientific contexts with basic objects
\[
    c_s,\qquad c_o,\qquad c_t,\qquad c_v,
\]
representing source, overlap, target, and validation regimes. The source context \(c_s\) is where the starting constellation is locally adequate. The target context \(c_t\) is where transport or extension is tested. The overlap context \(c_o\) is the common regime on which source-fitted and target-fitted charts are restricted and compared. The validation context \(c_v\) is held out from the selection obstruction and used only as a diagnostic. 

The intended global regime is probed by the finite cover
\[
    \{c_s,c_o,c_t\}.
\]
Thus a candidate constellation is not evaluated only by target fit. It must remain adequate on \(c_s\), fit \(c_t\), preserve the relevant source limit, and give compatible consequences on \(c_o\). The finite site is therefore the minimal structure used to test whether a representation still transports across regimes.

\subsection{Transition cards}

\begin{definition}[Transition card]
A transition card is a tuple
\[
    T =
    \bigl(
    \K_0,
    D_s,
    D_o,
    D_t,
    D_v,
    \{\Delta_j\}_{j=1}^{m}
    \bigr),
\]
where \(\K_0\) is the source constellation, \(D_s,D_o,D_t,D_v\) are observations in the source, overlap, target, and validation contexts, and
\(\{\Delta_j\}_{j=1}^{m}\) is a finite set of admissible representational moves. Each move is typed as a deformation or an extension.
\end{definition}

A transition card is the finite object on which the ranking problem is defined. It presents an artificial scientific agent with a source constellation \(\K_0\), evidence from several regimes, and a finite menu of possible ways to modify the source representation. The task is not to generate a theory from nothing, but to decide which candidate move best restores local-to-global coherence when the source constellation is tested outside its native regime. The evaluation label is not used by the ranking functional; it is used only after ranking to check whether the selected move is the intended deformation or extension.

The candidate move \(\Delta_j\) acts on the source constellation to produce
\[
    \K_j = \Delta_j(\K_0).
\]
This notation makes explicit that a move is an operation on the constellation. If \(\Delta_j\) is a deformation, then \(\K_j\) stays inside the original representational family: it changes parameters, correction terms, or law schemas while preserving the basic variables, constraints, and limiting interpretation. If \(\Delta_j\) is an extension, then \(\K_j\) belongs to an enlarged family: it adds a primitive, constraint, law schema, transformation rule, theoretical posit, or limit relation. The obstruction computation below is performed for each \(\K_j\), and the candidates are ranked by their resulting obstruction values.

The benchmark has three nested levels. A \emph{transition family} is an archetype of representational change, such as Galilean-to-Lorentzian velocity composition. A \emph{transition card} is one concrete instance within such a family, with its own observations and candidate moves. An \emph{observation record} is one local context-indexed numerical observation inside \(D_s,D_o,D_t\), or \(D_v\). Thus, statements such as ``40 source observations'' refer to 40 local records inside a single transition card, not to 40 distinct transition cards. For a schematic view of this hierarchy and a running example, see \ref{app:transition_card_anatomy}.

\subsection{Local fitting and restriction}

Each candidate constellation \(\K_j\) determines a model family
\[
    \mathcal M_j=\{f_{\K_j,\theta}:\theta\in\Theta_j\},
\]
where \(j\) indexes the candidate move, \(\theta\) denotes the parameters that can still be fitted inside that candidate, and \(f_{\K_j,\theta}\) is the predictive chart associated with \(\K_j\). In the Galilean-to-Lorentz case, for example, different values of \(j\) may correspond to the unchanged Galilean law, a bounded deformation, a wrong extension, or the Lorentzian extension; \(\theta\) then represents the adjustable parameters allowed within that chosen constellation. Fitting the candidate on a context-indexed dataset \(D\) selects the chart
\[
    \Fit(\K_j;D)
    =
    f_{\K_j,\widehat\theta(D)},
    \qquad
    \widehat\theta(D)
    =
    \argmin_{\theta\in\Theta_j}
    \sum_{(x,y)\in D}
    \ell\!\left(f_{\K_j,\theta}(x),y\right).
\]
The loss \(\ell\) depends on the observation type of the transition family; in the benchmark it is a normalized prediction loss over the observed quantities.

For each candidate \(\K_j\), the source and target fits are
\[
    \widehat{\K}_{j,s}=\Fit(\K_j;D_s),
    \qquad
    \widehat{\K}_{j,t}=\Fit(\K_j;D_t).
\]
The first subscript \(j\) identifies the candidate constellation, while \(s\) and \(t\) identify the source and target regimes. Thus \(\widehat{\K}_{j,s}\) is the version of candidate \(j\) fitted on source data, and \(\widehat{\K}_{j,t}\) is the same candidate fitted on target data. A global fit is also computed over the source, overlap, and target data: 
\[
    \widehat{\K}_{j,g}
    =
    \Fit(\K_j;D_s\cup D_o\cup D_t).
\]
The local fits test whether independently adapted source and target charts can agree on the overlap; the global fit supplies the residuals and structural
terms entering the obstruction functional.

Restriction evaluates the local charts on the overlap context:
\[
    \rho_{s\to o}(\widehat{\K}_{j,s}),
    \qquad
    \rho_{t\to o}(\widehat{\K}_{j,t}).
\]
Here \(s\to o\) and \(t\to o\) indicate that a source-fitted or target-fitted chart is being read on the common overlap regime. In the finite empirical setting, these restrictions are the predictions and constraint profiles induced by the fitted charts on the overlap observations \(D_o\). In the Galilean-to-Lorentz example, they ask how the low-velocity fit and the  higher-velocity fit behave on the same intermediate-velocity observations. If those restricted charts disagree, the candidate may fit locally but fail to transport coherently.

\subsection{Residual terms}

For each candidate \(\K_j\), the global fit \(\widehat{\K}_{j,g}\) is evaluated on the four context-indexed datasets. This gives normalized residuals
\[
    R_s(\K_j),\quad
    R_o(\K_j),\quad
    R_t(\K_j),\quad
    R_v(\K_j),
\]
where the subscripts denote source, overlap, target, and validation regimes. A typical residual has the form
\[
    R_c(\K_j)
    =
    \operatorname{nRMSE}
    \left(
        \widehat{\K}_{j,g}(D_c),
        D_c
    \right),
    \qquad
    c\in\{s,o,t,v\}.
\]
Thus \(R_s\) tests whether the candidate remains adequate in the original source regime, \(R_t\) tests whether it fits the new target regime, and \(R_o\) tests behavior in the intermediate overlap regime. In the Galilean-to-Lorentz example, these terms ask whether a candidate fits low-velocity data, higher subluminal data, and the intermediate regime where the two descriptions can be compared. The validation residual \(R_v\) is held out from the selection functional; it is used only for diagnostic reporting and for the broader obstruction signature used by the secondary kernel probe.

\subsection{Gluing residual}

The gluing test compares two local views of the same candidate constellation. For candidate \(\K_j\), the source fit \(\widehat{\K}_{j,s}\) and target fit \(\widehat{\K}_{j,t}\) are both evaluated on the overlap regime: 
\[
    \rho_{s\to o}(\widehat{\K}_{j,s})
    \simeq
    \rho_{t\to o}(\widehat{\K}_{j,t}).
\]
The symbol \(\simeq\) indicates approximate agreement rather than exact equality. In the finite benchmark, agreement means that the two restricted charts make compatible predictions and satisfy compatible constraint profiles on the same overlap observations.

We define the finite gluing residual as
\[
    G_{\mathrm{glue}}(\K_j)
    =
    d_o
    \left(
      \rho_{s\to o}(\widehat{\K}_{j,s}),
      \rho_{t\to o}(\widehat{\K}_{j,t})
    \right),
\]
where \(d_o\) is a normalized discrepancy on the overlap. In the Galilean-to-Lorentz example, this term compares what the low-velocity source-fitted chart and the higher-velocity target-fitted chart imply in the intermediate velocity regime. A small value means that the candidate transports smoothly across the overlap; a large value means that the local charts remain in tension even if each fits its own regime. 

This term is the finite counterpart of the sheaf gluing condition. A candidate may fit source and target data separately, yet fail to glue because the two fitted charts give incompatible consequences where their domains meet. This is the representational situation that the obstruction functional is designed to detect. Similar compatibility and consistency ideas appear in applied sheaf theory for sensor integration \citep{robinson2017}, cellular sheaf Laplacians and obstruction measures \citep{hansen_ghrist2019,ayzenberg2025}, and distributed systems or graph learning \citep{felber2025,bodnar2022}.

\begin{figure}[t]
\centering
\begin{tikzpicture}[
    node distance=1.0cm and 1.15cm,
    context/.style={rectangle, draw=black, rounded corners, align=center,
        minimum width=2.45cm, minimum height=0.78cm, font=\scriptsize, fill=gray!8},
    fitbox/.style={rectangle, draw=black, rounded corners, align=center,
        minimum width=2.25cm, minimum height=0.68cm, font=\scriptsize, fill=blue!6},
    term/.style={rectangle, draw=black, align=center,
        minimum width=2.1cm, minimum height=0.62cm, font=\scriptsize, fill=green!6},
    obs/.style={rectangle, draw=black, rounded corners, align=center,
        minimum width=2.7cm, minimum height=0.82cm, font=\scriptsize, fill=yellow!12},
    arrow/.style={-{Latex[length=1.8mm]}, thick},
    dashedarrow/.style={-{Latex[length=1.8mm]}, thick, dashed}
]

\node[context] (source) {Source context\\\(D_s\)};
\node[context, right=of source] (overlap) {Overlap context\\\(D_o\)};
\node[context, right=of overlap] (target) {Target context\\\(D_t\)};
\node[context, right=of target] (validation) {Validation\\\(D_v\)};

\node[fitbox, below=of source] (sfit) {Source fit\\\(\widehat{\K}_{j,s}\)};
\node[fitbox, below=of target] (tfit) {Target fit\\\(\widehat{\K}_{j,t}\)};
\node[fitbox, below=of overlap] (gfit) {Global fit\\\(\widehat{\K}_{j,g}\)};

\node[term, below=0.95cm of gfit] (glue) {Gluing residual\\\(G_{\mathrm{glue}}\)};
\node[term, left=of glue] (resid) {Residuals\\\(R_s,R_o,R_t\)};
\node[term, right=of glue] (struct) {Structure\\\(C_{\mathrm{viol}}\)\\\(P_{\mathrm{limit}},\Cost\)};
\node[obs, below=1.0cm of glue] (obstruction) {Selection obstruction\\\(\Obs_S(\K_j)\)};
\node[term, right=1.15cm of obstruction] (rv) {Diagnostic only\\\(R_v\)};

\draw[arrow] (source) -- (sfit);
\draw[arrow] (target) -- (tfit);
\draw[arrow] (source) -- (gfit);
\draw[arrow] (overlap) -- (gfit);
\draw[arrow] (target) -- (gfit);

\draw[dashedarrow] (sfit) -- node[above, sloped, font=\tiny]{restrict} (overlap);
\draw[dashedarrow] (tfit) -- node[above, sloped, font=\tiny]{restrict} (overlap);

\draw[arrow] (sfit) -- (glue);
\draw[arrow] (tfit) -- (glue);
\draw[arrow] (gfit) -- (resid);
\draw[arrow] (gfit) -- (struct);
\draw[arrow] (resid) -- (obstruction);
\draw[arrow] (glue) -- (obstruction);
\draw[arrow] (struct) -- (obstruction);
\draw[dashedarrow] (validation) |- (rv);
\end{tikzpicture}
\caption{Finite local-to-global obstruction computation. For each candidate constellation, source and target fits are restricted to the overlap to measure gluing failure. Global residuals, structural penalties, limit penalties, and representational cost define the selection obstruction \(\Obs_S\). The validation residual \(R_v\) is held out and used only for diagnostic signatures.}
\label{fig:constellation_site}
\end{figure}
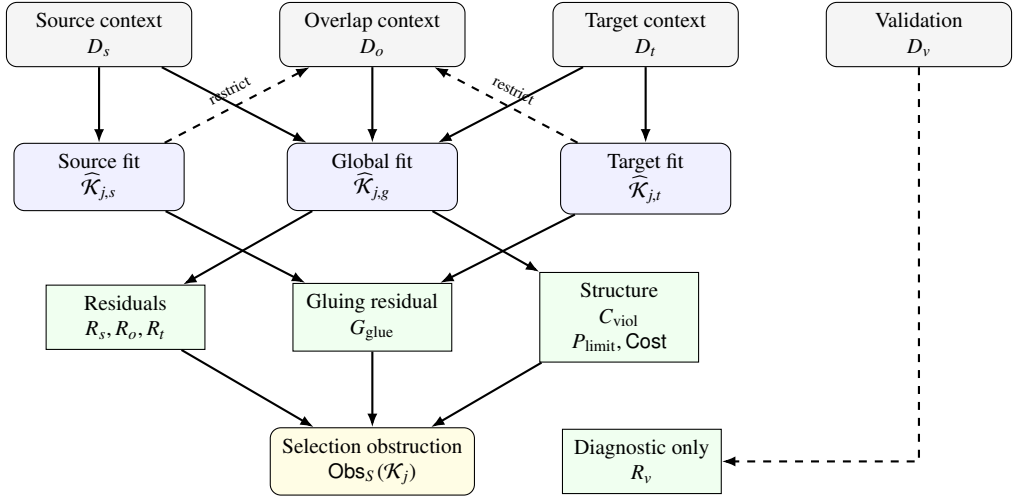

\subsection{Constraint and limit penalties}

Residual error alone is not enough. For a candidate constellation \(\K_j=\Delta_j(\K_0)\), good prediction on \(D_s,D_o,D_t\) does not guarantee that the candidate is an admissible scientific representation. It may fit the observations while violating an invariant, boundedness condition, conservation relation, monotonicity requirement, or source-regime limit. The obstruction functional therefore includes two structural penalties.

The constraint-violation term
\[
    C_{\mathrm{viol}}(\K_j)
\]
measures how strongly \(\K_j\) violates the admissibility constraints encoded in the transition family. These constraints are part of the representational constellation, not external afterthoughts: they specify what counts as an allowed description in the relevant contexts. In the benchmark, examples include the speed-bound constraint for Lorentzian velocity composition, finite-energy constraints for Planck-like radiation, low-density admissibility for virial corrections, and sign or monotonicity constraints for response functions.

The limit-preservation penalty
\[
    P_{\mathrm{limit}}(\K_j)
\]
measures whether \(\K_j\) recovers the source constellation \(\K_0\) in the appropriate source regime \(c_s\) or limiting domain. This term is essential because many successful extensions preserve older theories as limiting cases rather than simply replacing them. Relativistic kinetic energy must reduce to the Newtonian expression when \(v/c\ll 1\); Planck-like radiation must recover the appropriate classical behavior in its limiting regime; finite-angle pendulum dynamics must recover the small-angle approximation. In the Galilean-to-Lorentz case, an extension \(\K_j\in\mathcal F^{+}\) is acceptable only if it satisfies the target invariant-speed constraint while preserving the Galilean low-velocity limit of \(\K_0\).

\subsection{Cost of representational change}

Each candidate move \(\Delta_j\) is assigned a representational cost
\[
    \Cost(\Delta_j).
\]
This term penalizes changes of language that are not needed to restore coherence. A deformation usually has lower cost because it keeps the source constellation \(\K_0\) within the same representational family and modifies only a parameter, correction term, or law schema. An extension has higher cost when it adds a new primitive, constraint, transformation rule, theoretical posit, or limiting relation. The cost term prevents the ranking rule from selecting a more expressive constellation merely because it can fit the target data.

The cost is therefore not a generic complexity penalty. It is tied to the representational move being evaluated: the extension \(\K_j=\Delta_j(\K_0)\) must pay for the new resources it introduces. Such cost is justified only when it reduces residual, gluing, constraint, or limit obstruction enough to make the larger constellation more coherent than any admissible deformation. This follows the same general principle used in model selection and scientific representation: increased expressive power must be justified by improved coherence, not merely by local fit \citep{morgan_morrison1999,thagard2012,nersessian2008creating}.

\subsection{Obstruction functional}

The selection obstruction of candidate \(\K_j\) is
\begin{equation}
\begin{split}
    \Obs_S(\K_j)
    ={}& w_s R_s(\K_j)
      + w_o R_o(\K_j)
      + w_t R_t(\K_j)\\
     &+ w_g G_{\mathrm{glue}}(\K_j)
      + w_c C_{\mathrm{viol}}(\K_j)
      + w_l P_{\mathrm{limit}}(\K_j)
      + \lambda \Cost(\Delta_j).
\end{split}
\label{eq:obstruction}
\end{equation}
The subscript \(S\) marks this as the obstruction used for candidate selection.
It combines source, overlap, and target residuals with gluing failure,
constraint violation, limit failure, and representational cost. The held-out
validation residual \(R_v\) is not included in \(\Obs_S\); it is used only for
diagnostic reporting and for the broader obstruction signatures defined below.

A low-obstruction candidate is one with small source, overlap, and target
residuals \(R_s,R_o,R_t\), small gluing discrepancy
\(G_{\mathrm{glue}}\), low constraint and limit penalties
\(C_{\mathrm{viol}}\) and \(P_{\mathrm{limit}}\), and no unnecessary
representational cost \(\Cost(\Delta_j)\). A high-obstruction candidate may
fit one regime well, but fails as a coherent local-to-global representation
because one or more of these terms remains large. In the Galilean-to-Lorentz
case, for example, a fixed Galilean candidate may have low \(R_s\) but large
\(R_t\), \(C_{\mathrm{viol}}\), or \(G_{\mathrm{glue}}\). A Lorentzian
extension pays cost \(\Cost(\Delta_j)\), but is preferred only if the reduction
in residual, gluing, constraint, and limit penalties lowers the total
\(\Obs_S(\K_j)\).

The obstruction functional is the central quantitative object of the paper. It
turns the qualitative distinction between deformation and extension into a
ranking criterion: the selected move is the candidate with minimal
\(\Obs_S(\K_j)\).

\subsection{Transport, obstruction, and extension}

The obstruction ranking turns the transport-versus-extension distinction into a
decision rule. Let
\[
    j^\star=\argmin_{1\leq j\leq m}\Obs_S(\K_j)
\]
be the lowest-obstruction candidate for a transition card \(T\). The transition
is \emph{transportable} when the selected candidate \(\K_{j^\star}\) belongs to
the deformation family:
\[
    \K_{j^\star}\in\{\K_j:\Delta_j\ \text{is a deformation}\}.
\]
In this case, the source constellation remains adequate after bounded
within-language modification.

The transition is \emph{extension-required} when fixed-language deformations
retain higher obstruction and the selected candidate belongs to the extension
family:
\[
    \K_{j^\star}\in\{\K_j:\Delta_j\ \text{is an extension}\}.
\]
In this case, low obstruction is achieved only after enlarging the
representational language by adding a primitive, constraint, transformation
rule, law schema, theoretical posit, or limit relation. The Galilean-to-Lorentz
case has this form when Galilean deformations retain large \(R_t\),
\(G_{\mathrm{glue}}\), or \(C_{\mathrm{viol}}\), while the Lorentzian extension
lowers the total \(\Obs_S(\K_j)\) despite paying \(\Cost(\Delta_j)\).

This is the formal criterion used in the experiments. Discovery-like revision
is not identified with novelty alone. It is identified with a costed reduction
of local-to-global obstruction that cannot be achieved by deformation inside
the original constellation.

\subsection{Obstruction signatures}

For comparison across transition families, each candidate move is represented
by an obstruction signature
\[
    \Phi(T,\Delta_j)
    =
    \bigl[
    R_s(\K_j),
    R_o(\K_j),
    R_t(\K_j),
    R_v(\K_j),
    G_{\mathrm{glue}}(\K_j),
    C_{\mathrm{viol}}(\K_j),
    P_{\mathrm{limit}}(\K_j),
    \Cost(\Delta_j),
    \psi(G_{\K_j})
    \bigr],
\]
where \(\psi(G_{\K_j})\) denotes typed graph features of the candidate
constellation. The first entries record residual behavior across source,
overlap, target, and validation contexts; the next entries record gluing,
constraint, limit, and cost terms; and the final block records structural
features of the typed constellation graph. The validation residual \(R_v\)
appears here because the signature is used for analysis and kernel comparison,
not for candidate selection.

These signatures support two evaluations. First, their scalar projection through
Eq.~\eqref{eq:obstruction} gives the primary obstruction ranking by dropping
\(R_v\) and weighting the selection terms. Second, the full signature provides a
structured feature representation for the secondary representational
constellation kernel introduced in the next section.

\section{Constellation Kernels as a Representational Probe}
\label{sec:kernel}

The obstruction functional in Eq.~\eqref{eq:obstruction} is the primary
decision rule of the paper. It ranks candidate moves by local fit, gluing
compatibility, constraint satisfaction, limit preservation, and
representational cost. The kernel introduced here uses the resulting
obstruction signatures and constellation graphs to test whether these same
components induce a transferable similarity space across transition families.

Kernel methods are well suited to this comparison because the objects are
sparse, structured, and heterogeneous. Scientific transitions are structured
cases whose similarity depends on residual patterns, gluing behavior,
constraint profiles, limit preservation, and graph-level representational
commitments. General kernel methods compare such objects through feature maps
and inner products in implicit representation spaces
\citep{scholkopf_smola2002,shawe_taylor_cristianini2004}. Multiple-kernel and
block-kernel constructions allow distinct evidence sources to contribute
separate similarity components \citep{hofmann2008kernel}. Convolution kernels
compare structured objects by composing similarities over parts
\citep{haussler1999}, while graph kernels provide corresponding tools for
typed relational structures
\citep{borgwardt2005,shervashidze2011,vishwanathan2010}.

\subsection{Candidate signatures}

Each candidate move is represented as
\[
    a=(T,\Delta_j),
\]
where \(T\) is a transition card and \(\Delta_j\) is a candidate deformation or
extension. The candidate produces a constellation \(\K_j=\Delta_j(\K_0)\), and
its obstruction signature is
\[
    \PhiSig(T,\Delta_j)
    =
    \bigl[
    R_s(\K_j),\,
    R_o(\K_j),\,
    R_t(\K_j),\,
    R_v(\K_j),\,
    G_{\mathrm{glue}}(\K_j),\,
    C_{\mathrm{viol}}(\K_j),\,
    P_{\mathrm{limit}}(\K_j),\,
    \Cost(\Delta_j),\,
    \psi(G_{\K_j})
    \bigr],
\]
where \(\psi(G_{\K_j})\) denotes typed graph features of the candidate
constellation.

The signature separates the evidence used for selection from the information
used for comparison. Its weighted scalar projection, excluding the held-out
validation residual \(R_v\), gives the primary obstruction score
\(\Obs_S(\K_j)\). As a vector-valued representation, the full signature also
supports comparison across transition families. Two different physical
transitions may have analogous representational structure if both preserve the
source regime, fail under fixed-language deformation, exhibit gluing strain,
and become coherent only after introducing a new primitive, constraint, law
schema, or limiting relation.

\subsection{Graph features of a constellation}

The graph feature map \(\psi(G_{\K_j})\) summarizes the typed structure of a
candidate constellation. We write
\[
    \psi(G_{\K_j})
    =
    \bigl[
        n_V(G_{\K_j}),\,
        n_E(G_{\K_j}),\,
        n_3(G_{\K_j}),\,
        q(G_{\K_j})
    \bigr],
\]
where \(n_V\) gives typed node counts, \(n_E\) gives typed edge counts,
\(n_3\) gives typed triple counts, and \(q\) records representational
commitments. These commitments include \emph{invariant-speed structure},
\emph{low-speed limits}, \emph{quantization scales}, \emph{absolute time},
\emph{preferred frames}, \emph{limit relations}, \emph{removal of old posits},
and \emph{introduction of new constraints}.

The feature map distinguishes candidates whose numerical residuals may look
similar but whose constellation graphs differ. A deformation candidate
\(\K_{\theta}\) may alter a law-schema node \(\ell\) while preserving the
surrounding commitments of \(G_{\K_0}\). An extension candidate \(\K^{+}\) may
instead add new nodes or edges, such as
\((\ell,\mathrm{introduces},q)\),
\((\ell,\mathrm{constrains},c)\), or
\((\ell,\mathrm{preserves},r)\). In the Galilean-to-Lorentz case,
\(\psi(G_{\K_j})\) separates a deformation of the velocity-composition law from
an extension that introduces invariant-speed structure, Lorentz
transformations, and a low-speed limit relation.

The graph block \(\psi(G_{\K_j})\) enters the kernel through
\(k_{\mathrm{graph}}\), not through the primary selection obstruction
\(\Obs_S(\K_j)\). Thus the primary ranking remains the obstruction ranking from
Eq.~\eqref{eq:obstruction}, while the graph features test whether structural
changes in \(G_{\K_j}\) help compare candidate moves across transition
families. The typed-count representation can be replaced by random-walk,
shortest-path, Weisfeiler--Lehman, or other graph-kernel constructions
\citep{gartner2003,borgwardt2005,shervashidze2011,vishwanathan2010}.

\subsection{Additive block kernel}

The obstruction signature is block structured. For a candidate row
\(a=(T,\Delta_j)\), let
\[
    \PhiSig(a)
    =
    \bigl[
        z_{\mathrm{res}}(a),\,
        z_{\mathrm{glue}}(a),\,
        z_{\mathrm{con}}(a),\,
        z_{\mathrm{lim}}(a),\,
        \psi(G_{\K_j})
    \bigr],
\]
where \(z_{\mathrm{res}}\) contains standardized residual features,
\(z_{\mathrm{glue}}\) the standardized gluing feature,
\(z_{\mathrm{con}}\) the standardized constraint feature,
\(z_{\mathrm{lim}}\) the standardized limit feature, and
\(\psi(G_{\K_j})\) the typed graph features. Each block measures a different
aspect of representational adequacy: fit, overlap compatibility,
admissibility, source-limit preservation, and graph-level representational
structure.

The kernel compares two candidate rows \(a=(T,\Delta_i)\) and
\(b=(T',\Delta_j)\) by adding the similarities of these blocks:
\begin{equation}
    k(a,b)
    =
    \alpha_{\mathrm{res}} k_{\mathrm{res}}(a,b)
    + \alpha_{\mathrm{glue}} k_{\mathrm{glue}}(a,b)
    + \alpha_{\mathrm{con}} k_{\mathrm{con}}(a,b)
    + \alpha_{\mathrm{lim}} k_{\mathrm{lim}}(a,b)
    + \alpha_{\mathrm{graph}} k_{\mathrm{graph}}(a,b).
\label{eq:kernel}
\end{equation}
The coefficients \(\alpha_{\mathrm{res}},\alpha_{\mathrm{glue}},
\alpha_{\mathrm{con}},\alpha_{\mathrm{lim}},\alpha_{\mathrm{graph}}\) control
how much each evidence block contributes to similarity. Unlike
\(\Obs_S(\K_j)\), which is a scalar ranking score for one transition card,
\(k(a,b)\) compares two candidate moves across cards or families.

For a numeric block
\(B\in\{\mathrm{res},\mathrm{glue},\mathrm{con},\mathrm{lim}\}\), we use a
radial-basis kernel over standardized block vectors:
\[
    k_B(a,b)
    =
    \exp\!\left(
    -\frac{\|z_B(a)-z_B(b)\|^2}{2\sigma_B^2}
    \right).
\]
Thus \(k_{\mathrm{res}}\) compares the residual profiles of two candidates,
\(k_{\mathrm{glue}}\) compares their overlap-compatibility behavior,
\(k_{\mathrm{con}}\) compares their constraint-violation profiles, and
\(k_{\mathrm{lim}}\) compares their source-limit behavior. The graph block uses
a normalized linear kernel over typed constellation features:
\[
    k_{\mathrm{graph}}(a,b)
    =
    \frac{
        \langle \psi(G_{\K_a}),\psi(G_{\K_b})\rangle
    }{
        \|\psi(G_{\K_a})\|\,\|\psi(G_{\K_b})\|+\varepsilon
    }.
\]
Here \(G_{\K_a}\) and \(G_{\K_b}\) are the typed constellation graphs associated
with the two candidate moves. The inner product is large when the candidates
share representational commitments, for example when both graphs contain
features for a new constraint, a preserved limit relation, or a newly introduced
theoretical primitive.

The additive form mirrors the structure of the obstruction signature. The terms
\(k_{\mathrm{res}}, k_{\mathrm{glue}}, k_{\mathrm{con}}, k_{\mathrm{lim}}\),
and \(k_{\mathrm{graph}}\) keep fit, overlap compatibility, admissibility,
limit preservation, and graph-level commitments as separate evidence sources,
while the weights \(\alpha_{\mathrm{res}},\alpha_{\mathrm{glue}},
\alpha_{\mathrm{con}},\alpha_{\mathrm{lim}},\alpha_{\mathrm{graph}}\) combine
them into one similarity measure. The kernel therefore tests whether the same
objects used to compute obstruction also define a useful geometry of
representational moves across transition families.

\subsection{Kernel ranking task}

For each transition card \(T\), the ranking task is to order its candidate
moves \(\{\Delta_j\}_{j=1}^m\). In the kernel experiment, each candidate row
\(a=(T,\Delta_j)\) is represented by its obstruction signature
\(\PhiSig(a)\). Given signatures from training families, the kernel \(k(a,b)\)
defines similarities among candidate moves, and a kernel scoring model assigns
a score to each held-out candidate. Candidates inside the held-out card are then
ranked by this score.

The evaluation uses a leave-family-out protocol. If \(f\) is the held-out
transition family, signatures from all other families are used for training,
and candidates from \(f\) are used only for evaluation. This tests whether the
signature blocks \(z_{\mathrm{res}},z_{\mathrm{glue}},z_{\mathrm{con}},
z_{\mathrm{lim}}\), together with \(\psi(G_{\K_j})\), carry structure that
transfers across transition types rather than merely describing one family.

The kernel task remains secondary to direct obstruction ranking. Direct
obstruction ranking asks whether the theory-defined functional
\(\Obs_S(\K_j)\) selects the intended move within each transition card. Kernel
ranking asks whether the resulting signatures define a useful representational
geometry over candidate moves. Thus the kernel is a probe of the structure
induced by the obstruction components, not a replacement for the obstruction
criterion itself.

\subsection{Workflow}

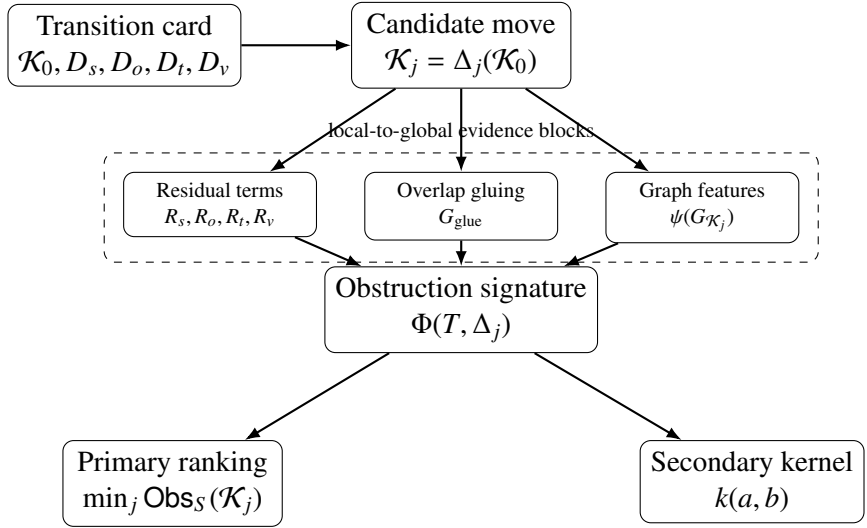
\begin{figure}[t]
\centering
\begin{tikzpicture}[
    node distance=1.25cm and 1.45cm,
    box/.style={
        rectangle,
        rounded corners,
        draw=black,
        align=center,
        minimum width=2.9cm,
        minimum height=0.9cm,
        font=\small
    },
    smallbox/.style={
        rectangle,
        rounded corners,
        draw=black,
        align=center,
        minimum width=2.55cm,
        minimum height=0.8cm,
        font=\scriptsize
    },
    arrow/.style={-{Latex[length=2mm]}, thick}
]

\node[box] (card) {Transition card\\
\(\K_0,D_s,D_o,D_t,D_v\)};
\node[box, right=of card] (cands) {Candidate move\\
\(\K_j=\Delta_j(\K_0)\)};

\node[smallbox, below left=1.1cm and 0.45cm of cands] (fit)
{Residual terms\\\(R_s,R_o,R_t,R_v\)};
\node[smallbox, below=1.1cm of cands] (restrict)
{Overlap gluing\\\(G_{\mathrm{glue}}\)};
\node[smallbox, below right=1.1cm and 0.45cm of cands] (graph)
{Graph features\\\(\psi(G_{\K_j})\)};

\node[box, below=2.35cm of cands] (signature) {Obstruction signature\\
\(\Phi(T,\Delta_j)\)};

\node[box, below left=1.15cm and 0.55cm of signature] (obsrank) {Primary ranking\\
\(\min_j \Obs_S(\K_j)\)};
\node[box, below right=1.15cm and 0.55cm of signature] (kernel) {Secondary kernel\\
\(k(a,b)\)};

\draw[arrow] (card) -- (cands);
\draw[arrow] (cands) -- (fit);
\draw[arrow] (cands) -- (restrict);
\draw[arrow] (cands) -- (graph);

\draw[arrow] (fit) -- (signature);
\draw[arrow] (restrict) -- (signature);
\draw[arrow] (graph) -- (signature);

\draw[arrow] (signature) -- (obsrank);
\draw[arrow] (signature) -- (kernel);

\node[
    draw=black,
    dashed,
    rounded corners,
    fit=(fit)(restrict)(graph),
    inner sep=0.25cm,
    label={[font=\scriptsize]above:local-to-global evidence blocks}
] {};
\end{tikzpicture}
\caption{Workflow from transition cards to obstruction ranking and constellation kernels. Each candidate move \(\Delta_j\) produces a constellation \(\K_j\), whose residual, gluing, structural, and graph features form the obstruction signature \(\Phi(T,\Delta_j)\). The scalar obstruction \(\Obs_S\) gives the primary decision rule, while the full signature defines the secondary kernel \(k(a,b)\) for probing representational similarity across transition families.}
\label{fig:kernel_workflow}
\end{figure}

Figure~\ref{fig:kernel_workflow} summarizes how the primary obstruction
ranking and the secondary kernel probe use the same candidate-level evidence.
A transition card \(T\) supplies the source constellation \(\K_0\), context
data \(D_s,D_o,D_t,D_v\), and candidate moves \(\Delta_j\). Each move produces
a candidate constellation \(\K_j=\Delta_j(\K_0)\). From \(\K_j\), the finite
local-to-global computation extracts residual terms, overlap-restriction and
gluing terms, structural penalties, and typed graph features. These components
form the obstruction signature \(\Phi(T,\Delta_j)\). The scalar projection
\(\Obs_S(\K_j)\) gives the primary ranking, while the full signature defines
the secondary kernel \(k(a,b)\) for comparing candidate moves across transition
families.

\section{Experimental Design}
\label{sec:experiments}

The experimental design evaluates the ranking rule defined by
\(\Obs_S(\K_j)\): for each transition card \(T\), does the lowest-obstruction
candidate correspond to a deformation inside the source constellation or to an
extension of that constellation? Each card supplies a source regime \(D_s\), an
overlap regime \(D_o\), a target regime \(D_t\), and candidate moves
\(\{\Delta_j\}_{j=1}^m\). The source regime tests whether \(\K_0\) remains
locally adequate; the overlap regime tests whether independently fitted charts
restrict compatibly; and the target regime tests whether low obstruction can be
achieved without adding a new primitive, constraint, law schema, transformation
rule, or limiting relation. The use of controlled physical transition families
follows the computational scientific-discovery tradition of testing structure
recovery on interpretable scientific systems \citep{langley1987scientific}.
Modern equation-discovery and sparse-dynamics methods continue this emphasis on
recovering meaningful structure rather than only minimizing curve-fitting error
\citep{schmidt2009distilling,brunton2016sindy}. Recent symbolic-regression and
AI-for-science work further sharpens this setting by evaluating whether learned
expressions capture the right structural form, not merely predictive accuracy
\citep{udrescu2020ai_feynman,cranmer2020symbolic}.

\subsection{Transition families}

The benchmark contains six physics-inspired transition families. Each family
defines a source constellation \(\K_0\), context regimes
\((D_s,D_o,D_t,D_v)\), and a finite candidate set
\(\{\Delta_j\}_{j=1}^m\). Three families are
\emph{deformation-sufficient}: some within-language move \(\Delta_j\) produces
low obstruction without changing the representational resources of \(\K_0\).
Three families are \emph{extension-required}: all admissible deformations retain
high obstruction, and low \(\Obs_S(\K_j)\) requires an enlarged constellation
\(\K_j=\Delta_j(\K_0)\).

The deformation-sufficient families are
\[
\begin{array}{ll}
\text{small-angle pendulum} & \longrightarrow \text{finite-angle correction},\\
\text{ideal gas law} & \longrightarrow \text{virial correction},\\
\text{Ohm's law} & \longrightarrow \text{temperature-dependent resistance}.
\end{array}
\]
The extension-required families are
\[
\begin{array}{ll}
\text{Galilean velocity composition} & \longrightarrow \text{Lorentzian velocity composition},\\
\text{Newtonian kinetic energy} & \longrightarrow \text{relativistic kinetic energy},\\
\text{Rayleigh--Jeans radiation} & \longrightarrow \text{Planck-like blackbody law}.
\end{array}
\]
For each family, the source, overlap, target, and validation regimes are fixed
explicitly, and the intended representational move is used only for evaluation.
Table~\ref{tab:benchmark_definition} summarizes the transition types, regimes,
and key structural constraints.

\begin{table}[t]
\centering
\footnotesize
\caption{Benchmark transition families. Each row defines a source-to-target regime sequence and the structural constraint or limiting relation used to test whether the source constellation transports or requires extension.}
\label{tab:benchmark_definition}
\setlength{\tabcolsep}{4pt}
\begin{tabularx}{\linewidth}{@{}L{0.20\linewidth}L{0.14\linewidth}L{0.23\linewidth}L{0.15\linewidth}Y@{}}
\toprule
Family & Type & Source-to-target regimes & Intended move & Key structure \\
\midrule
Galilean \(\rightarrow\) Lorentz velocity
& Extension
& Low \(\to\) intermediate \(\to\) high subluminal velocities
& Lorentz composition
& Invariant speed; low-speed limit \\
Newtonian \(\rightarrow\) relativistic energy
& Extension
& Low \(\to\) moderate \(\to\) relativistic \(v/c\)
& Relativistic energy
& Domain \(v<c\); monotonicity; Newtonian limit \\
Rayleigh--Jeans \(\rightarrow\) Planck radiation
& Extension
& Long \(\to\) intermediate \(\to\) shorter wavelength
& Planck-like law
& Finite energy; low-frequency limit \\
Small-angle \(\rightarrow\) finite pendulum
& Deformation
& Small \(\to\) moderate \(\to\) finite angle
& Finite-angle correction
& Small-angle limit \\
Ideal gas \(\rightarrow\) virial equation
& Deformation
& Low \(\to\) moderate \(\to\) higher density
& Virial correction
& Low-density limit; monotonicity \\
Ohm \(\rightarrow\) temperature resistance
& Deformation
& Reference \(\to\) wider \(\to\) broad temperature range
& Temperature-dependent \(R\)
& Reference-temperature limit \\
\bottomrule
\end{tabularx}
\end{table}

\subsection{Contexts and observations}

Each transition card contains four context-indexed datasets
\(D_s,D_o,D_t,D_v\), corresponding to source, overlap, target, and validation
regimes. The source data \(D_s\) test whether the initial constellation
\(\K_0\) remains adequate in its native regime. The overlap data \(D_o\) are
the common regime on which \(\widehat{\K}_{j,s}\) and
\(\widehat{\K}_{j,t}\) are restricted and compared. The target data \(D_t\)
test whether candidate \(\K_j=\Delta_j(\K_0)\) transports or requires
extension. The validation data \(D_v\) are excluded from \(\Obs_S\) and used
only as a held-out diagnostic.

An observation record is one local datum \((x,y,c)\), where \(x\) denotes the
input variables, \(y\) the observed quantity, and
\(c\in\{s,o,t,v\}\) the context label. Each record may also activate
context-specific constraints used in \(C_{\mathrm{viol}}(\K_j)\) or limiting
checks used in \(P_{\mathrm{limit}}(\K_j)\). Thus the evaluation is not only
residual fitting through \(R_s,R_o,R_t\). A candidate must preserve the source
regime, agree on the overlap through \(G_{\mathrm{glue}}\), satisfy structural
constraints through \(C_{\mathrm{viol}}\), preserve relevant limits through
\(P_{\mathrm{limit}}\), and fit the target regime. Table~\ref{tab:transition_card_information}
summarizes how each card component contributes to this local-to-global test.

\begin{table}[t]
\centering
\footnotesize
\caption{Information contained in a transition card. Each component contributes to the finite local-to-global test by exposing a distinct failure mode.}
\label{tab:transition_card_information}
\setlength{\tabcolsep}{4pt}
\begin{tabularx}{\linewidth}{@{}L{0.22\linewidth}YY@{}}
\toprule
Component & Role in the local-to-global test & Failure mode captured \\
\midrule
Source constellation
& Starting representational language and commitments.
& Source language not preserved. \\
Source observations \(D_s\)
& Native regime where \(\K_0\) should remain adequate.
& High \(R_s\). \\
Overlap observations \(D_o\)
& Common regime for comparing restricted source- and target-fitted charts.
& High \(G_{\mathrm{glue}}\) or \(R_o\). \\
Target observations \(D_t\)
& New regime where transport or extension is tested.
& High \(R_t\). \\
Validation observations \(D_v\)
& Held-out regime for diagnostic checking.
& High \(R_v\). \\
Constraints and limits
& Encode admissibility, invariants, and limiting behavior.
& High \(C_{\mathrm{viol}}\) or \(P_{\mathrm{limit}}\). \\
Candidate deformations/extensions
& Competing moves \(\Delta_j\) applied to \(\K_0\).
& Incorrect minimum-obstruction move. \\
\bottomrule
\end{tabularx}
\end{table}

\subsection{Candidate classes}

For each transition card \(T\), the benchmark supplies a finite set of candidate
moves \(\{\Delta_j\}_{j=1}^m\). Each move acts on the source constellation to
produce \(\K_j=\Delta_j(\K_0)\), and each move has a role: \emph{base},
\emph{deformation}, \emph{incorrect alternative}, or \emph{intended move}. The
\emph{base} move leaves \(\K_0\) unchanged and tests whether the source
constellation already transports. A \emph{deformation} changes a law schema,
parameterization, or correction term while remaining inside the original
representational language. An \emph{incorrect alternative} is a controlled
distractor: it may add flexibility or an extension-like resource, but it does
not remove the relevant local-to-global obstruction. The \emph{intended move}
is the benchmark-correct deformation or extension used only for evaluation.

This candidate set defines the ranking problem. The obstruction functional uses
only the generated constellations \(\K_j\), their residuals
\(R_s,R_o,R_t\), gluing discrepancy \(G_{\mathrm{glue}}\), structural penalties
\(C_{\mathrm{viol}},P_{\mathrm{limit}}\), and representational cost
\(\Cost(\Delta_j)\). It does not use the intended label when ranking. The label
is used afterward to determine whether the minimum-obstruction candidate
\[
    j^\star=\argmin_j \Obs_S(\K_j)
\]
matches the benchmark-correct move and whether its transition type is
deformation or extension.

The candidate structure separates three diagnostic questions. First, does the
base move keep obstruction low, so that \(\K_0\) transports without change?
Second, does some deformation \(\K_{\theta}\) reduce obstruction while
preserving the source language? Third, if low obstruction is achieved only by
an extension \(\K^{+}\), does that extension introduce the structural resource
needed to restore gluing, constraints, and limits rather than merely adding
flexibility? The present experiment evaluates this obstruction-based ranking
problem; autonomous generation of \(\Delta_j\) by symbolic search, program
synthesis, or language-model proposal is left for future work.

\subsection{Obstruction weights and representational costs}

The selection obstruction \(\Obs_S\) uses fixed weights for the seven terms in
Eq.~\eqref{eq:obstruction}:
\[
R_s(\K_j),\quad R_o(\K_j),\quad R_t(\K_j),\quad
G_{\mathrm{glue}}(\K_j),\quad
C_{\mathrm{viol}}(\K_j),\quad
P_{\mathrm{limit}}(\K_j),\quad
\Cost(\Delta_j).
\]
These weights are not learned from the evaluation labels. They set the relative
scale of the controlled benchmark: \(w_s\) and \(w_o\) give unit weight to
source and overlap fit, \(w_t\) and \(w_g\) give higher weight to target fit and
gluing, \(w_c\) gives the strongest weight to structural admissibility, and
\(\lambda\) keeps representational cost active but smaller than the main
coherence terms. Sensitivity analysis in Section~\ref{sec:results} tests
whether the main conclusions depend on narrow tuning of these values.

\begin{table}[t]
\centering
\small
\caption{Reference weights in the selection obstruction \(\Obs_S\). The values define the fixed scoring rule used before sensitivity analysis.}
\label{tab:obstruction_weights}
\begin{tabular}{lcc}
\toprule
Evidence term & Symbol & Weight \\
\midrule
Source residual & \(w_s\) & 1.00 \\
Overlap residual & \(w_o\) & 1.00 \\
Target residual & \(w_t\) & 1.50 \\
Gluing residual & \(w_g\) & 1.50 \\
Constraint violation & \(w_c\) & 2.00 \\
Limit penalty & \(w_l\) & 1.50 \\
Representational cost & \(\lambda\) & 0.25 \\
\bottomrule
\end{tabular}
\end{table}

Representational cost \(\Cost(\Delta_j)\) is attached to the move that produces
\(\K_j=\Delta_j(\K_0)\), not to the fitted residuals
\(R_s,R_o,R_t\). The base move has cost \(0\). Deformations receive small
positive costs, typically \(0.4\)--\(0.6\), because they preserve the source
language while changing a parameterization, correction term, or law schema.
Extensions receive larger costs because they add a new primitive, constraint,
transformation rule, law schema, theoretical posit, or limiting relation. In
the benchmark, intended extensions have costs around \(1.5\)--\(1.7\), while
controlled incorrect extensions are assigned comparable extension-like costs
when they add representational capacity without the intended structural role.
The weighted term \(\lambda\Cost(\Delta_j)\) therefore penalizes unnecessary
language change while still allowing an extension to win when it sufficiently
reduces \(R_t(\K_j)\), \(G_{\mathrm{glue}}(\K_j)\),
\(C_{\mathrm{viol}}(\K_j)\), or \(P_{\mathrm{limit}}(\K_j)\).

\subsection{Primary evaluation: obstruction ranking}

The primary evaluation ranks candidate moves by the selection obstruction
\(\Obs_S\). For each transition card
\(T=(\K_0,D_s,D_o,D_t,D_v,\{\Delta_j\}_{j=1}^m)\), each candidate move produces
\[
    \K_j=\Delta_j(\K_0).
\]
The score \(\Obs_S(\K_j)\) is then computed from Eq.~\eqref{eq:obstruction}
using the source, overlap, and target residuals \(R_s,R_o,R_t\), the gluing
term \(G_{\mathrm{glue}}\), the structural penalties
\(C_{\mathrm{viol}},P_{\mathrm{limit}}\), and the weighted cost
\(\lambda\Cost(\Delta_j)\). Candidates are ranked in ascending obstruction, and
the selected move is
\[
    \widehat j(T)
    =
    \argmin_{1\leq j\leq m} \Obs_S(\K_j).
\]

The intended move is not used in the ranking. It is used only after selection
to evaluate whether \(\widehat j(T)\) matches the benchmark-correct candidate.
We report top-1 accuracy, mean reciprocal rank, and transition-type accuracy.
Top-1 accuracy asks whether the selected candidate is exactly correct; mean
reciprocal rank measures how high the intended candidate appears in the
obstruction ranking; and transition-type accuracy asks whether the selected move
has the correct type, deformation for deformation-sufficient families and
extension for extension-required families.

\subsection{Secondary evaluation: constellation-kernel ranking}

The secondary evaluation tests whether the obstruction signatures define a
transferable representation space across transition families. Each candidate
row \(a=(T,\Delta_j)\) is represented by
\[
    \PhiSig(T,\Delta_j),
\]
which contains residual terms \(R_s,R_o,R_t,R_v\), gluing discrepancy
\(G_{\mathrm{glue}}\), structural penalties
\(C_{\mathrm{viol}},P_{\mathrm{limit}}\), cost \(\Cost(\Delta_j)\), and graph
features \(\psi(G_{\K_j})\). Candidate rows are compared by the additive
constellation kernel \(k(a,b)\) in Eq.~\eqref{eq:kernel}.

We use a leave-one-family-out protocol. For a held-out transition family \(f\),
candidate signatures from the other five families are used to score and rank
the candidates in \(f\). This tests whether the signature blocks
\(z_{\mathrm{res}},z_{\mathrm{glue}},z_{\mathrm{con}},z_{\mathrm{lim}}\) and
\(\psi(G_{\K_j})\) transfer across transition types rather than only separating
candidates within one family. Leave-one-group-out evaluation is a standard way
to assess generalization across structured groups rather than across
exchangeable individual samples
\citep{stone1974crossvalidatory,hastie2009elements}.

The kernel experiment remains secondary to direct obstruction ranking. The
primary question is whether \(\Obs_S(\K_j)\) selects the intended move inside
each transition card. The kernel asks a different question: whether the
candidate signatures induced by the obstruction terms and constellation graphs
form a useful similarity geometry across families.

\subsection{Stress tests and robustness}

Two additional analyses test the stability of the obstruction ranking. The
stress test expands the candidate set
\(\{\Delta_j\}_{j=1}^m\) with additional incorrect formulas, randomized
candidates, and matched-cost incorrect extensions. For each expanded card, we
recompute \(\Obs_S(\K_j)\) and compare the intended candidate with the best
incorrect or matched-cost alternative. This tests whether the intended move is
selected because it reduces local-to-global obstruction, rather than because it
has favorable cost \(\Cost(\Delta_j)\) or merely adds flexibility.

The robustness analysis perturbs the context-indexed observations
\(D_s,D_o,D_t,D_v\). We add noise to the observation values and reduce the
number of available records, then recompute the obstruction ranking. The goal is
not to simulate every possible experimental uncertainty, but to test whether the
selected move \(\widehat j(T)\) remains stable when the finite evidence is
degraded. In this way, the robustness sweep probes whether the diagnosis
depends mainly on coherent local-to-global structure or on fragile sampling
details.

\subsection{Experimental questions}

The experiments are organized around the finite theory-shift detection problem:
given a source constellation \(\K_0\), candidate moves \(\{\Delta_j\}\), and
context-indexed evidence \(D_s,D_o,D_t,D_v\), can obstruction ranking detect
whether the source representation transports or requires extension? The
experimental blocks test the ranking itself, the deformation-versus-extension
classification, the contribution of obstruction components, and the stability
of the diagnosis under controlled perturbations. Table~\ref{tab:experimental_questions}
summarizes this structure.

\begin{table}[H]
\centering
\small
\caption{Experimental questions addressed by the benchmark. The experiments test whether finite obstruction ranking detects transport versus extension and whether the resulting diagnosis is stable under controlled perturbations.}
\label{tab:experimental_questions}
\begin{tabular}{p{0.31\linewidth}p{0.41\linewidth}p{0.20\linewidth}}
\toprule
Question & Diagnostic target & Main evidence \\
\midrule
Does obstruction ranking select the intended move?
& Whether the benchmark-correct candidate has the lowest \(\Obs_S(\K_j)\).
& Primary ranking metrics; candidate landscapes. \\

Does the selected move have the correct transition type?
& Whether deformation-sufficient families select deformations and extension-required families select extensions.
& Transition-type accuracy; extension/deformation examples. \\

Does obstruction improve on target fit alone?
& Whether \(G_{\mathrm{glue}}\), \(C_{\mathrm{viol}}\), \(P_{\mathrm{limit}}\), and \(\Cost\) add information beyond \(R_t\).
& Baselines and ablations; margin ledger. \\

Which terms explain the ranking?
& How residual, gluing, structural, limit, and cost terms shape the margin between the intended candidate and its nearest competitor.
& Obstruction-margin ledger; candidate landscapes. \\

Is the diagnosis stable?
& Whether \(\widehat j(T)\) is robust to weight changes, expanded incorrect alternatives, noise, and reduced record availability.
& Sensitivity, stress, and robustness figures. \\

Does the kernel add a secondary representation probe?
& Whether \(\PhiSig(T,\Delta_j)\) and \(\psi(G_{\K_j})\) induce transferable similarity across transition families.
& Combined kernel probe; appendix kernel tables. \\
\bottomrule
\end{tabular}
\end{table}

The table also fixes the role of the secondary analyses. Direct obstruction
ranking is the decision rule. The kernel asks whether the same signatures define
a useful similarity geometry across families. Stress and robustness analyses
test where the obstruction diagnosis remains stable and where boundary cases
appear.

\subsection{Summary of experimental logic}

The experimental construction applies the finite obstruction test to a
collection of transition cards
\[
    T=(\K_0,D_s,D_o,D_t,D_v,\{\Delta_j\}_{j=1}^m).
\]
For each candidate move \(\Delta_j\), the card generates a candidate
constellation \(\K_j=\Delta_j(\K_0)\). The candidate is fitted in source and
target regimes, restricted to the overlap, evaluated by residual, gluing,
constraint, limit, and cost terms, and ranked by \(\Obs_S(\K_j)\). Thus the
core computation is
\[
    (T,\Delta_j)
    \;\mapsto\;
    \K_j
    \;\mapsto\;
    \bigl(
        R_s,R_o,R_t,
        G_{\mathrm{glue}},
        C_{\mathrm{viol}},
        P_{\mathrm{limit}},
        \Cost
    \bigr)
    \;\mapsto\;
    \Obs_S(\K_j).
\]
A deformation-sufficient family is successful when a deformation candidate has
the lowest obstruction. An extension-required family is successful when
fixed-language candidates remain obstructed and the intended extension is
selected. The benchmark therefore tests the central diagnostic claim of the
paper: transport is adequate when coherence can be restored inside the original
constellation, while extension is required when coherence appears only after
enlarging the representational language.

Algorithms~\ref{alg:transition_card_construction} and
\ref{alg:obstruction_ranking} give the two core operations used in the
experiments. Additional protocols for the constellation kernel, stress tests,
and robustness sweeps are collected in \ref{app:auxiliary_protocols}.

\begin{algorithm}[H]
\caption{Transition-card construction}
\label{alg:transition_card_construction}
\KwInput{Transition families \(\mathcal F\); context regimes
\((D_s,D_o,D_t,D_v)\); admissible move classes \(\mathcal D\)}
\KwOutput{Transition-card collection \(\mathcal T\)}
\(\mathcal T \leftarrow \emptyset\)\;
\ForEach{transition family \(f\in\mathcal F\)}{
  Specify the source constellation \(\K_0^{(f)}\)\;
  Generate context-indexed datasets
  \(D_s^{(f)},D_o^{(f)},D_t^{(f)},D_v^{(f)}\) for the source,
  overlap, target, and validation regimes\;
  Define admissible candidate moves
  \[
      \Delta^{(f)}
      =
      \{\Delta^{(f)}_1,\ldots,\Delta^{(f)}_{m_f}\},
  \]
  including the base move, deformations, controlled incorrect alternatives,
  and intended moves\;
  Form the transition card
  \[
      T^{(f)}
      =
      \bigl(
      \K_0^{(f)},
      D_s^{(f)},D_o^{(f)},D_t^{(f)},D_v^{(f)},
      \Delta^{(f)}
      \bigr)
  \]\;
  \(\mathcal T \leftarrow \mathcal T\cup\{T^{(f)}\}\)\;
}
\Return{\(\mathcal T\)}
\end{algorithm}

Algorithm~\ref{alg:transition_card_construction} defines the finite card
collection. Algorithm~\ref{alg:obstruction_ranking} ranks the candidate moves
inside each card by their selection obstruction.

\begin{algorithm}[H]
\caption{Finite obstruction ranking}
\label{alg:obstruction_ranking}
\KwInput{Transition card
\(T=(\K_0,D_s,D_o,D_t,D_v,\{\Delta_j\}_{j=1}^m)\)}
\KwOutput{Selected move \(\widehat{\Delta}(T)\), ranked moves, and obstruction signatures}
\ForEach{candidate move \(\Delta_j\)}{
  Form the candidate constellation \(\K_j=\Delta_j(\K_0)\)\;
  Fit local charts
  \(\widehat{\K}_{j,s}=\Fit(\K_j;D_s)\) and
  \(\widehat{\K}_{j,t}=\Fit(\K_j;D_t)\)\;
  Restrict local charts to the overlap:
  \(\rho_{s\to o}(\widehat{\K}_{j,s})\) and
  \(\rho_{t\to o}(\widehat{\K}_{j,t})\)\;
  Compute residuals \(R_s(\K_j),R_o(\K_j),R_t(\K_j)\) for selection
  and \(R_v(\K_j)\) for diagnostics\;
  Compute gluing residual
  \[
      G_{\mathrm{glue}}(\K_j)
      =
      d_o\!\left(
      \rho_{s\to o}(\widehat{\K}_{j,s}),
      \rho_{t\to o}(\widehat{\K}_{j,t})
      \right)
  \]\;
  Compute structural and cost terms
  \(C_{\mathrm{viol}}(\K_j)\), \(P_{\mathrm{limit}}(\K_j)\), and
  \(\Cost(\Delta_j)\)\;
  Evaluate \(\Obs_S(\K_j)\) using Eq.~\eqref{eq:obstruction}\;
  Store the signature \(\PhiSig(T,\Delta_j)\)\;
}
Rank moves so that
\[
    \Obs_S(\K_{(1)})
    \leq
    \Obs_S(\K_{(2)})
    \leq
    \cdots
    \leq
    \Obs_S(\K_{(m)}),
    \qquad
    \K_{(r)}=\Delta_{(r)}(\K_0).
\]\;
\(\widehat{\Delta}(T)\leftarrow\Delta_{(1)}\)\;
\Return{\(\widehat{\Delta}(T)\), ranked moves, and \(\{\PhiSig(T,\Delta_j)\}_{j=1}^m\)}
\end{algorithm}

\FloatBarrier

\section{Results}
\label{sec:results}

The results evaluate whether the selection obstruction \(\Obs_S(\K_j)\)
detects when a source constellation transports by deformation and when it
requires extension. The benchmark contains 30 transition cards across six
families: three deformation-sufficient families and three extension-required
families. For each card \(T\), candidate moves \(\Delta_j\) generate
constellations \(\K_j=\Delta_j(\K_0)\), and the candidates are ranked by
ascending \(\Obs_S(\K_j)\). The intended candidate and transition type are used
only for evaluation.

Three findings organize the section. First, direct obstruction ranking selects
the intended deformation or extension in most cards and perfectly separates
deformation-sufficient from extension-required transition types in this
benchmark. Second, baselines, ablations, and stress tests show that the result
is not explained by target residual alone, representational cost, or added
flexibility. Third, the constellation kernel gives a useful secondary probe of
the representation space, but remains weaker than direct obstruction ranking
and is not the decision rule.

\subsection{Transition families and candidate sets}

Table~\ref{tab:families_counts} summarizes the evaluated cards. Each transition
family contributes five cards. Deformation-sufficient families require a
bounded move inside the source representational language. Extension-required
families require a new primitive, constraint, law schema, transformation rule,
or limiting relation. Each card contains the unchanged source constellation,
candidate deformations, controlled incorrect alternatives, and the intended
deformation or extension.

\begin{table}[t]
\centering
\footnotesize
\caption{Transition families, transition type, number of transition cards, candidate count, and intended representational move. Family labels are shortened for readability.}
\label{tab:families_counts}
\setlength{\tabcolsep}{4pt}
\begin{tabularx}{\linewidth}{@{}L{0.22\linewidth}L{0.15\linewidth}ccY@{}}
\toprule
Family & Type & \shortstack{Transition\\cards} & Candidates & Intended move \\
\midrule
Galilean \(\rightarrow\) Lorentz & extension & 5 & 5 & Lorentz composition with invariant-speed structure \\
Newtonian \(\rightarrow\) relativistic & extension & 5 & 4 & Relativistic kinetic-energy schema \\
Rayleigh--Jeans \(\rightarrow\) Planck & extension & 5 & 4 & Quantization-scale extension \\
Small-angle \(\rightarrow\) finite & deformation & 5 & 3 & Finite-angle deformation \\
Ideal gas \(\rightarrow\) virial & deformation & 5 & 4 & Quadratic virial deformation \\
Ohm \(\rightarrow\) temperature & deformation & 5 & 3 & Temperature-dependent resistance deformation \\
\bottomrule
\end{tabularx}
\end{table}

\begin{figure}[t]
\centering
\includegraphics[width=0.92\linewidth]{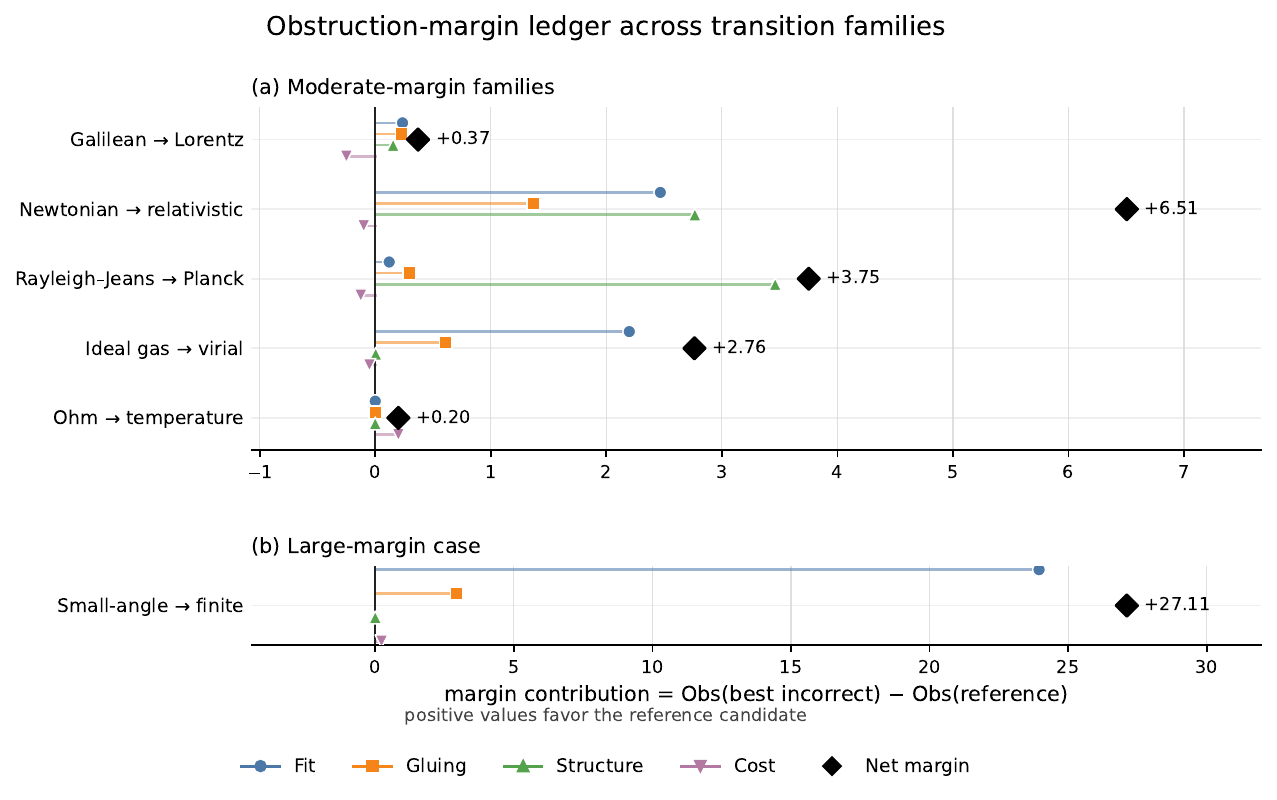}
\caption{Obstruction-margin ledger across transition families. Each row shows one representative transition card. The reference candidate is the benchmark-correct deformation or extension, and the best incorrect alternative is the lowest-obstruction non-reference candidate. Markers show signed contributions of fit, gluing, structure, and cost to the margin \(\Obs_S(\mathrm{best\ incorrect})-\Obs_S(\mathrm{reference})\); black diamonds mark the net margin. Positive values favor the reference candidate. The ledger shows which obstruction components support the benchmark-correct move before the detailed candidate landscapes in Figure~\ref{fig:running_example_candidate_landscapes}.}
\label{fig:obstruction_margin_ledger}
\end{figure}

Figure~\ref{fig:obstruction_margin_ledger} shows how obstruction components
support or oppose the intended move. For one representative card from each
family, the figure compares the benchmark-correct reference candidate with the
best incorrect alternative. The signed margin decomposes into fit, gluing,
structural, and cost contributions, so the reader can see whether the ranking is
driven by residuals, overlap compatibility, constraints and limits, or
representational cost.

\subsection{Obstruction ranking identifies the intended move}

The primary evaluation uses no supervised training. For each transition card,
the selected candidate is
\[
    \widehat j(T)=\argmin_{1\leq j\leq m}\Obs_S(\K_j).
\]
As shown in Table~\ref{tab:obstruction_metrics}, obstruction ranking selects
the intended candidate in 27 of 30 cards, with mean reciprocal rank \(0.950\).
It also achieves perfect transition-type accuracy: selected candidates are
deformations in deformation-sufficient families and extensions in
extension-required families.

This is the main empirical evidence for the proposed diagnosis. The score
\(\Obs_S\) combines source, overlap, and target residuals \(R_s,R_o,R_t\),
gluing discrepancy \(G_{\mathrm{glue}}\), constraint and limit penalties
\(C_{\mathrm{viol}},P_{\mathrm{limit}}\), and cost
\(\lambda\Cost(\Delta_j)\). The metrics in Table~\ref{tab:obstruction_metrics}
measure three aspects of the ranking. \emph{Top-1} is the fraction of cards for
which the lowest-obstruction candidate is the benchmark-correct move.
\emph{Mean reciprocal rank} is the average of \(1/r(T)\), where \(r(T)\) is the
rank position of the benchmark-correct candidate in card \(T\). \emph{Type
accuracy} is the fraction of cards for which the selected move has the correct
transition type: deformation for deformation-sufficient families and extension
for extension-required families. The perfect transition-type accuracy indicates
that the obstruction terms organize the selected move as transport or extension,
not merely as a low-error fit.

\begin{table}[t]
\centering
\caption{Primary obstruction-ranking performance. Top-1 measures exact
candidate selection, MRR measures the rank of the benchmark-correct candidate,
and type accuracy measures whether the selected move has the correct
deformation-versus-extension type.}
\label{tab:obstruction_metrics}
\begin{tabular}{lcccc}
\toprule
Evaluation & Cases & Top-1 & MRR & Type accuracy \\
\midrule
Obstruction ranking & 30 & 0.900 & 0.950 & 1.000 \\
Stress-expanded candidate set & 30 & 0.900 & 0.925 & 1.000 \\
\bottomrule
\end{tabular}
\end{table}

\begin{figure}[t]
\centering
\includegraphics[width=0.92\linewidth]{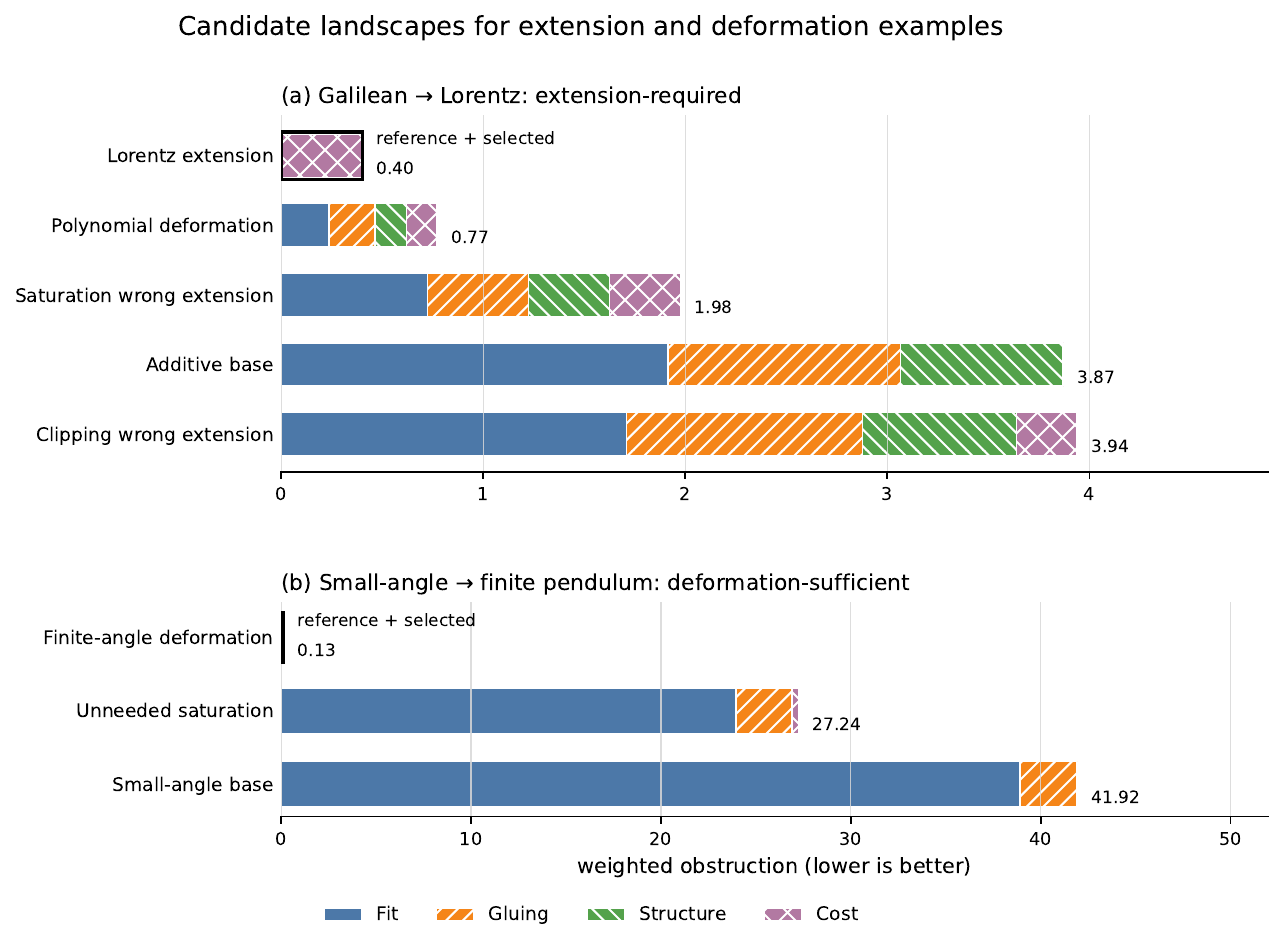}
\caption{Candidate landscapes for one extension-required and one deformation-sufficient transition card. Bars show total weighted obstruction decomposed into fit, gluing, structure, and cost contributions. Lower total obstruction is better. Panel~(a) shows a Galilean-to-Lorentz card in which the Lorentzian extension is the benchmark-correct and selected candidate. Panel~(b) shows a small-angle-to-finite-pendulum card in which the finite-angle deformation is benchmark-correct and selected. The examples illustrate the distinction between paying representational cost for a necessary extension and restoring coherence by deformation inside the original language.}
\label{fig:running_example_candidate_landscapes}
\end{figure}

Figure~\ref{fig:running_example_candidate_landscapes} shows two representative
candidate landscapes. In the Galilean-to-Lorentz card, the Lorentzian extension
pays representational cost but lowers residual, gluing, and structural
obstruction enough to become the selected candidate. In the
small-angle-to-finite-pendulum card, a finite-angle deformation restores
coherence without enlarging the representational language.

\subsection{Baselines and direct obstruction ablations}

The baseline comparisons test whether the theory-shift diagnosis is already
captured by residual error or by a simpler costed score. Each row in
Table~\ref{tab:direct_baselines} defines an alternative scalar ranking score
\(S_{\mathrm{base}}(\K_j)\); candidates are ranked by ascending
\(S_{\mathrm{base}}\), exactly as candidates are ranked by ascending
\(\Obs_S\) in the main evaluation. The target-only baseline uses only
\(R_t(\K_j)\). The source--target baseline uses \(R_s(\K_j)+R_t(\K_j)\). The
source--overlap--target baseline uses \(R_s(\K_j)+R_o(\K_j)+R_t(\K_j)\).
The residual--cost baseline adds \(\lambda\Cost(\Delta_j)\) to the residual
terms. The residual--gluing baseline adds \(G_{\mathrm{glue}}(\K_j)\) but
omits constraint, limit, and cost terms. The full obstruction row is
Eq.~\eqref{eq:obstruction}.

Table~\ref{tab:direct_baselines} shows that target residual alone matches full
obstruction in aggregate top-1 accuracy, but not in transition-type accuracy.
Residual-only scores can often find a plausible candidate, but they do not
reliably organize the selected move as deformation or extension. Adding cost to
residuals performs worse because it can favor cheaper but structurally
inadequate candidates. Adding gluing restores perfect transition-type accuracy.

\begin{table}[t]
\centering
\small
\caption{Direct non-kernel baselines for candidate ranking. Each row defines a
scalar score used to rank candidates by ascending value. The comparison tests
whether the full obstruction functional adds diagnostic structure beyond
residual fit, cost, or gluing alone.}
\label{tab:direct_baselines}
\begin{tabular}{lccc}
\toprule
Ranking score & Top-1 & MRR & Type accuracy \\
\midrule
Target residual only & 0.900 & 0.950 & 0.900 \\
Source + target residual & 0.833 & 0.917 & 0.867 \\
Source + overlap + target residual & 0.900 & 0.950 & 0.900 \\
Residual + cost & 0.667 & 0.833 & 0.833 \\
Residual + gluing & 0.900 & 0.950 & 1.000 \\
Full obstruction & 0.900 & 0.950 & 1.000 \\
\bottomrule
\end{tabular}
\end{table}

The added value of obstruction is therefore diagnostic rather than simply
aggregate predictive accuracy. The local-to-global terms expose why a candidate
wins: whether it fits the target, agrees on the overlap, preserves limits,
satisfies constraints, and pays justified representational cost. In this
benchmark, the full obstruction functional and residual-only ranking can agree
on top-1 counts, but they differ in their ability to preserve the
transport-versus-extension structure of the task.

Table~\ref{tab:direct_obstruction_ablation} reports direct ablations of
Eq.~\eqref{eq:obstruction}. In each ``No \(X\)'' row, candidates are reranked
after setting the corresponding obstruction term to zero while leaving the
remaining terms unchanged. For example, ``No gluing'' removes
\(w_gG_{\mathrm{glue}}(\K_j)\), ``No limit'' removes
\(w_lP_{\mathrm{limit}}(\K_j)\), and ``No cost'' removes
\(\lambda\Cost(\Delta_j)\). The residual-only and residual-plus-cost rows are
included again to make the comparison with the baseline scores explicit.

Removing the limit term reduces both top-1 and transition-type accuracy,
confirming the role of source-regime preservation. Removing gluing increases
aggregate top-1 in this finite benchmark but lowers transition-type accuracy,
mainly because it avoids penalizing noisy virial cases. This result should not
be read as evidence against gluing. It shows that \(G_{\mathrm{glue}}\) is a
structural term: it enforces the local-to-global interpretation of a move as
transport or extension, while also revealing where finite data and weights
create strain. Removing constraints or cost has little aggregate effect in the
original 30-card ranking, although \(C_{\mathrm{viol}}\) and \(\Cost\) remain
important for interpreting the selected moves and for the kernel probe below.

\begin{table}[t]
\centering
\small
\caption{Direct obstruction ablations. Each row reranks candidates after
removing one term from Eq.~\eqref{eq:obstruction}, without using the kernel.
``No gluing'', for example, removes \(w_gG_{\mathrm{glue}}\), while ``No
limit'' removes \(w_lP_{\mathrm{limit}}\).}
\label{tab:direct_obstruction_ablation}
\begin{tabular}{lccc}
\toprule
Ablation score & Top-1 & MRR & Type accuracy \\
\midrule
Full obstruction & 0.900 & 0.950 & 1.000 \\
No source residual & 0.900 & 0.933 & 1.000 \\
No overlap residual & 0.900 & 0.950 & 1.000 \\
No target residual & 0.900 & 0.950 & 1.000 \\
No gluing & 0.967 & 0.983 & 0.967 \\
No constraints & 0.900 & 0.950 & 1.000 \\
No limit & 0.867 & 0.933 & 0.967 \\
No cost & 0.900 & 0.950 & 1.000 \\
Residual only & 0.900 & 0.950 & 0.900 \\
Residual + cost & 0.667 & 0.833 & 0.833 \\
\bottomrule
\end{tabular}
\end{table}

\subsection{Weight and cost sensitivity}

\begin{figure}[t]
\centering
\includegraphics[width=0.92\linewidth]{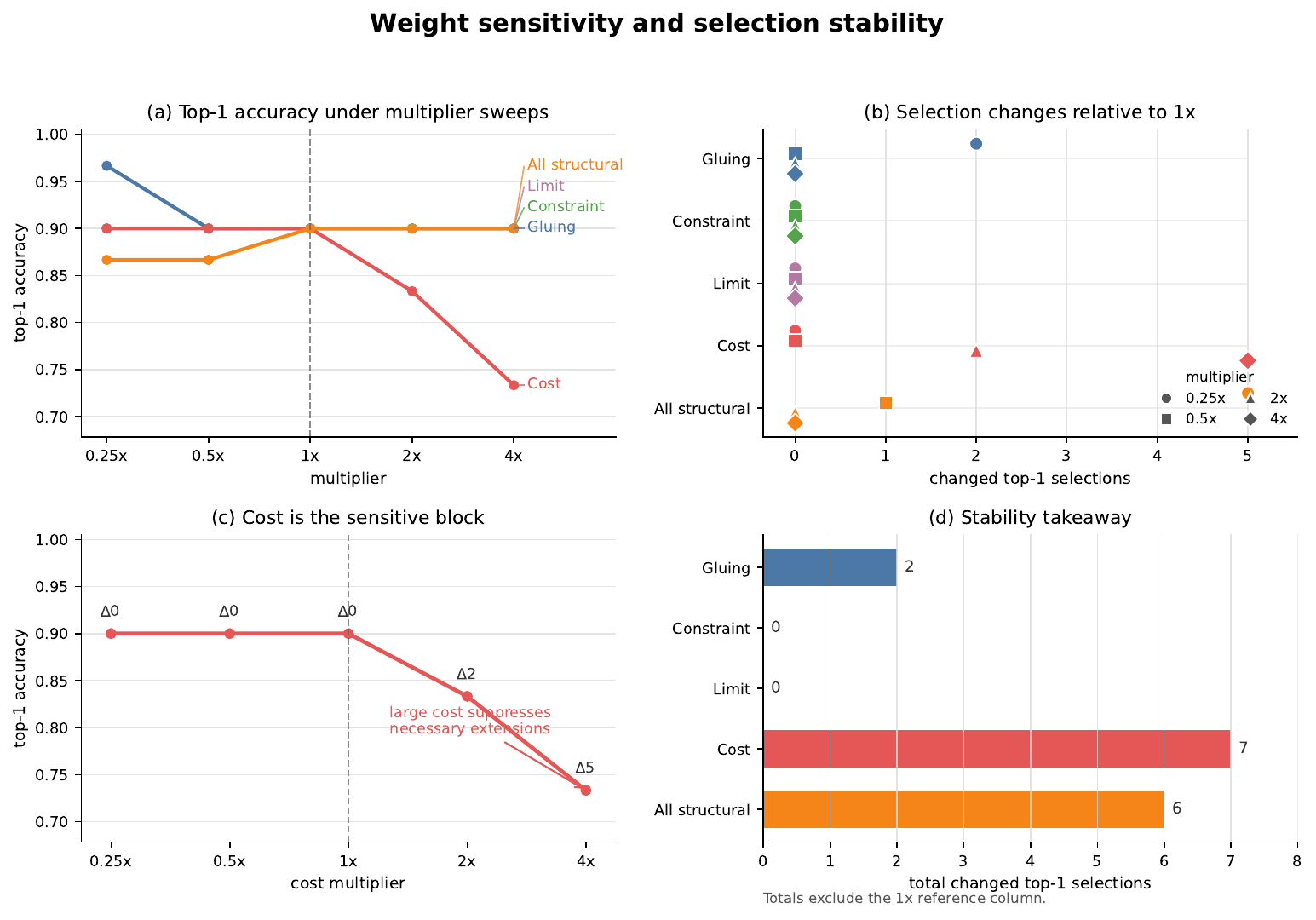}
\caption{Weight sensitivity and selection stability. Panel~(a) shows top-1 accuracy as each obstruction block is multiplied relative to the reference \(1\times\) setting. Panel~(b) shows the number of selected moves \(\widehat j(T)\) that change relative to the reference ranking. Panel~(c) isolates the cost multiplier, the most sensitive block: large cost values reduce accuracy by over-penalizing necessary extensions. Panel~(d) summarizes selection instability across blocks. Moderate perturbations are stable, while excessive cost penalization is the main failure mode.}
\label{fig:weight_sensitivity}
\end{figure}

Figure~\ref{fig:weight_sensitivity} summarizes sensitivity to the weights in
Eq.~\eqref{eq:obstruction}. The reference setting is the \(1\times\) multiplier.
Each sweep multiplies one obstruction block while leaving the other blocks fixed
at their reference values, then recomputes the selected move
\(\widehat j(T)=\argmin_j\Obs_S(\K_j)\). Moderate changes to residual, gluing,
constraint, and limit blocks leave top-1 accuracy and selected candidates
largely unchanged. The cost multiplier is the most sensitive block: excessive
cost penalization suppresses necessary extensions and changes several top-1
selections. This is the expected failure mode for a costed theory-shift
criterion. Cost should discourage unnecessary language change, but not prevent
an extension when it sharply lowers \(R_t\), \(G_{\mathrm{glue}}\),
\(C_{\mathrm{viol}}\), or \(P_{\mathrm{limit}}\).

Held-out validation residuals \(R_v\) are not used in \(\Obs_S\), but they check
whether low-obstruction choices remain plausible outside the fitted source,
overlap, and target regimes. Across the 30 cards, intended candidates have mean
validation residual \(0.037\), selected top-1 candidates have mean validation
residual \(0.045\), best incorrect candidates have mean validation residual
\(0.252\), and source-base candidates have mean validation residual \(0.982\).
Low obstruction therefore generally transfers to held-out regimes, with the
main exceptions occurring in the virial boundary cases discussed next.

\subsection{Wrong-candidate stress test}

The stress test asks whether the intended move wins because it removes
local-to-global obstruction, rather than merely because it has more expressive
capacity. For each transition card \(T\), the original candidate set
\(\{\Delta_j\}_{j=1}^m\) is expanded with additional incorrect formulas,
randomized formula perturbations, and matched-cost incorrect extensions. Each
new move \(\Delta_j\) produces a candidate constellation
\(\K_j=\Delta_j(\K_0)\), and all candidates are reranked by
\(\Obs_S(\K_j)\). The ranking remains stable: top-1 accuracy stays at
\(0.900\), mean reciprocal rank remains high at \(0.925\), and no matched-cost
incorrect extension beats the intended extension.

\begin{figure}[t]
\centering
\includegraphics[width=0.92\linewidth]{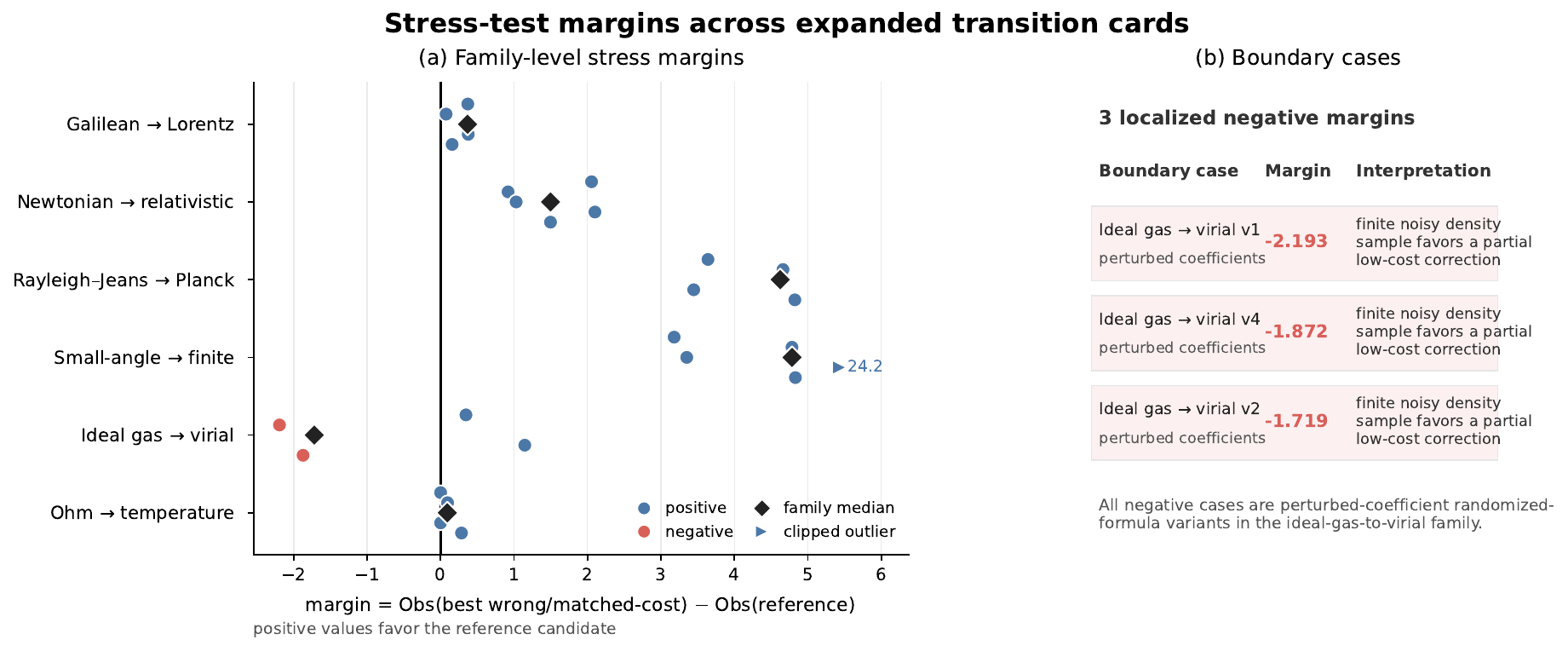}
\caption{Stress-test margins across expanded transition cards. Panel~(a) shows, for each transition family, the margin \(M(T)=\Obs_S(\K_{\mathrm{best\ incorrect}})-\Obs_S(\K_{\mathrm{ref}})\) between the best incorrect or matched-cost alternative and the benchmark-correct reference candidate. Positive margins indicate that the reference candidate retains lower obstruction; negative margins indicate boundary cases where a non-reference alternative becomes preferred. Panel~(b) highlights the negative cases, which are concentrated in virial/randomized-formula variants. These cases mark informative benchmark boundaries rather than a general collapse of the obstruction criterion.}
\label{fig:stress_margins}
\end{figure}

Figure~\ref{fig:stress_margins} reports the margin
\[
    M(T)
    =
    \Obs_S(\K_{\mathrm{best\ incorrect}})
    -
    \Obs_S(\K_{\mathrm{ref}}),
\]
where \(\K_{\mathrm{ref}}\) is the benchmark-correct candidate and
\(\K_{\mathrm{best\ incorrect}}\) is the lowest-obstruction non-reference or
matched-cost alternative in the expanded set. Positive margins mean that the
reference candidate still has lower obstruction. Negative margins mark cases
where a non-reference candidate becomes preferred.

Most transition cards retain positive margins. The negative margins are
concentrated in perturbed-coefficient ideal-gas-to-virial variants, where the
ranking selects a lower-cost linear virial deformation or a randomized
perturbed-coefficient formula instead of the intended quadratic virial
deformation. In formal terms, those alternatives reduce enough of
\(G_{\mathrm{glue}}(\K_j)\), \(R_o(\K_j)\), or \(\Cost(\Delta_j)\) on finite
noisy density samples to offset their weaker structural interpretation. These
cases identify a boundary of the finite benchmark: partial low-cost corrections
can appear more coherent than the intended quadratic correction when the
available density evidence is noisy or sparse.

\subsection{Robustness to noise and reduced evidence}

The robustness sweep perturbs the context-indexed datasets
\(D_s,D_o,D_t,D_v\) and recomputes the selected move
\[
    \widehat j(T)=\argmin_j \Obs_S(\K_j).
\]
For each noise level \(\eta\) and retained-record fraction \(q\), observation
values are perturbed and only a fraction \(q\) of records is retained in each
context. The figure reports the resulting mean top-1 accuracy over transition
cards.

\begin{figure}[t]
\centering
\includegraphics[width=0.92\linewidth]{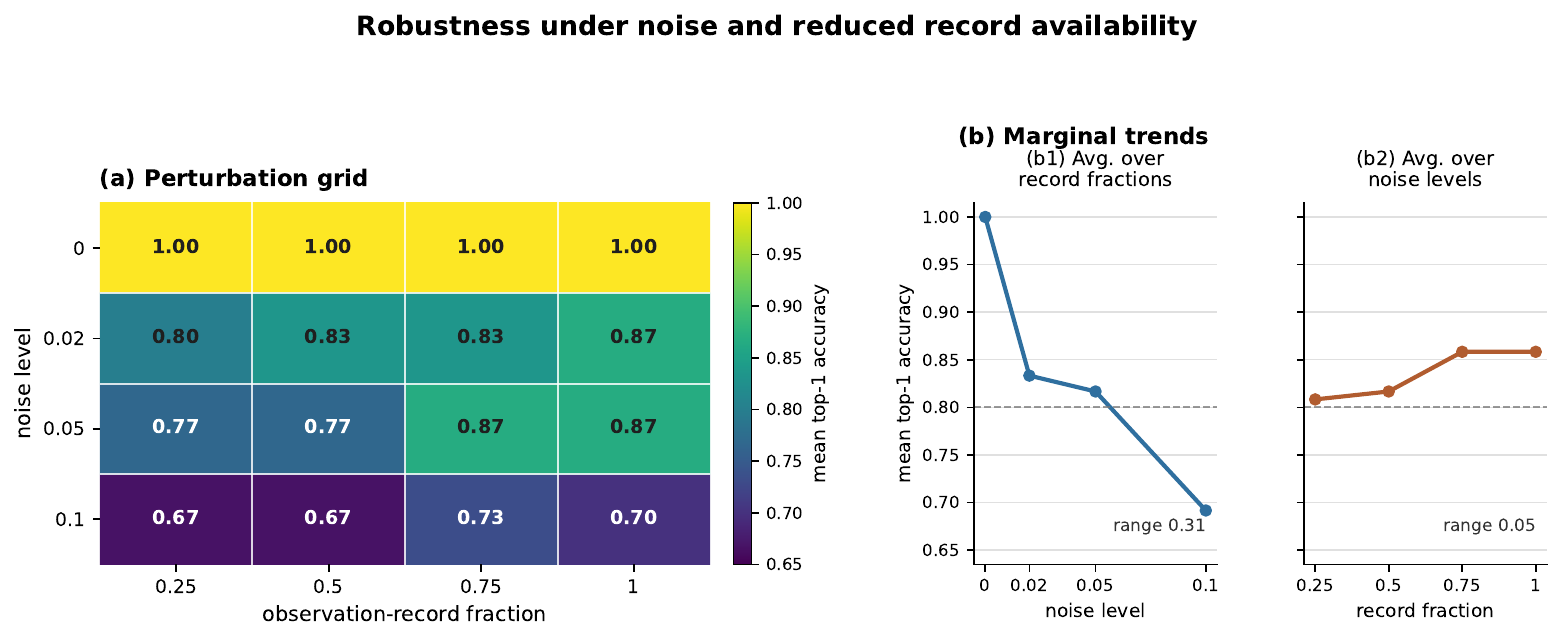}
\caption{Robustness under observation noise and reduced record availability. Panel~(a) reports mean top-1 accuracy after recomputing \(\widehat j(T)=\argmin_j\Obs_S(\K_j)\) for each noise level \(\eta\) and retained-record fraction \(q\). Panels~(b1) and~(b2) summarize the same grid by averaging over record fractions and noise levels, respectively. Accuracy decreases mainly as observation noise increases, while moderate reductions in record availability have a weaker effect.}
\label{fig:robustness_heatmap}
\end{figure}

Figure~\ref{fig:robustness_heatmap} shows the full perturbation grid and the
marginal trends. Accuracy remains high under moderate subsampling, meaning that
the obstruction ranking does not require all observation records to recover the
same transport-versus-extension diagnosis. Accuracy decreases more sharply as
noise increases, because noise directly corrupts the residuals
\(R_s,R_o,R_t\) and the overlap comparison entering
\(G_{\mathrm{glue}}\). In this benchmark, the diagnosis is therefore more
sensitive to noisy local evidence than to moderate reductions in record
availability.

\subsection{Secondary constellation-kernel probe}

The constellation kernel evaluates whether the candidate signatures form a
transferable representation space across transition families. Each candidate
row \(a=(T,\Delta_j)\) is represented by the obstruction signature
\(\PhiSig(T,\Delta_j)\), including residual, gluing, constraint, limit, cost,
and graph-feature blocks. The kernel \(k(a,b)\) then compares candidate moves
through these blocks rather than through \(\Obs_S(\K_j)\) alone. In
leave-one-family-out evaluation, the kernel achieves top-1 accuracy \(0.600\),
mean reciprocal rank \(0.783\), and transition-type accuracy \(0.800\). These
values are below direct obstruction ranking, but they show that
\(\PhiSig(T,\Delta_j)\) and \(\psi(G_{\K_j})\) carry cross-family structure.

The family-level results are heterogeneous. The kernel performs strongly on
Rayleigh--Jeans \(\rightarrow\) Planck and Newtonian \(\rightarrow\)
relativistic energy, but weakly on Ohm \(\rightarrow\) temperature resistance
and some deformation-sufficient families. This is expected in the small-sample
setting: the kernel must infer similarity across families from few structured
examples, whereas the selection obstruction \(\Obs_S(\K_j)\) evaluates each
candidate directly inside its own transition card.

\begin{figure}[t]
\centering
\includegraphics[width=\linewidth]{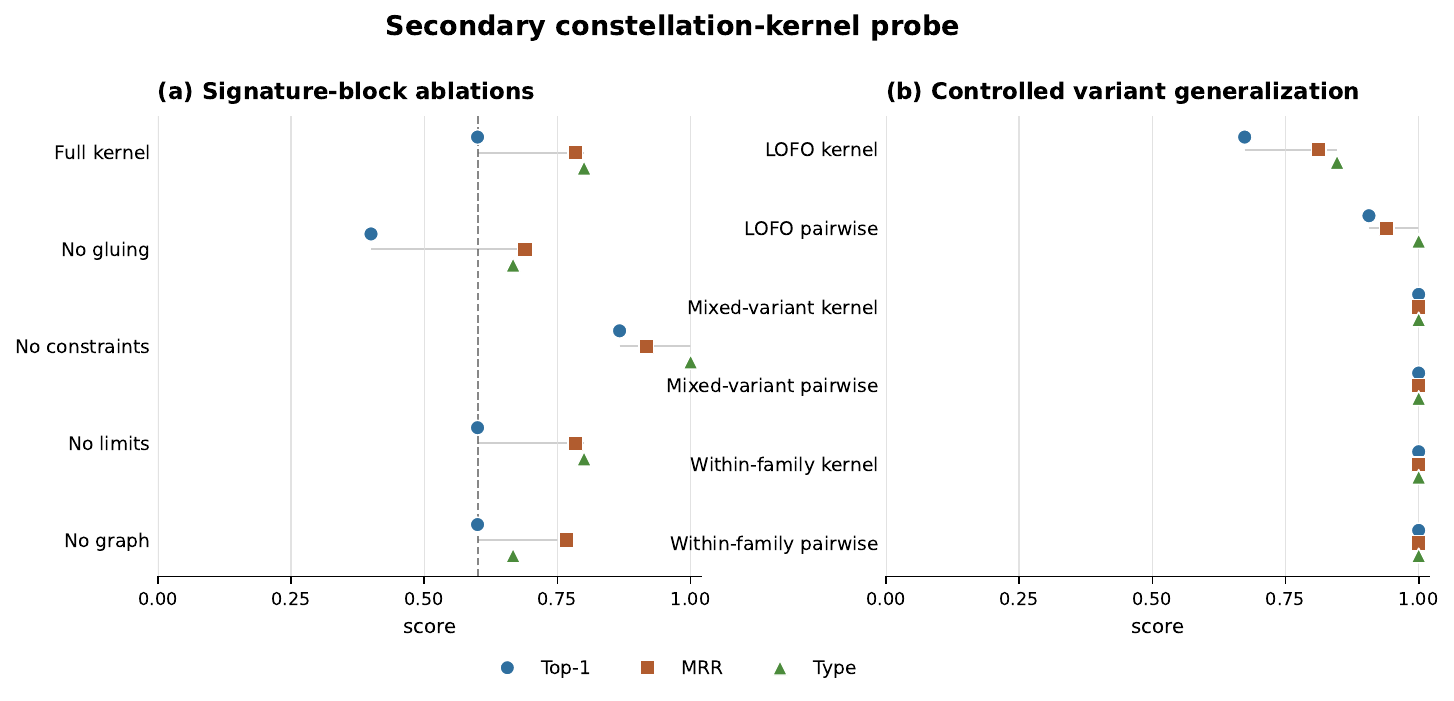}
\caption{Secondary constellation-kernel probe. Panel~(a) ablates the kernel blocks in Eq.~\eqref{eq:kernel}. Removing \(k_{\mathrm{glue}}\) reduces ranking and type discrimination, while removing \(k_{\mathrm{graph}}\) mainly affects transition-type prediction; the ``no constraints'' result indicates that \(k_{\mathrm{con}}\) requires better calibration for kernel generalization. Panel~(b) compares controlled generalization protocols: within-family and mixed-variant generalization are nearly saturated, while leave-one-family-out remains the harder analogical-transfer setting. The kernel probes the geometry of \(\PhiSig(T,\Delta_j)\) and \(\psi(G_{\K_j})\); it does not replace direct obstruction ranking by \(\Obs_S(\K_j)\).}
\label{fig:kernel_probe_summary}
\end{figure}

Figure~\ref{fig:kernel_probe_summary} summarizes two diagnostics. Panel~(a)
ablates the kernel blocks in Eq.~\eqref{eq:kernel}. Removing the gluing block
\(k_{\mathrm{glue}}\) reduces ranking and transition-type discrimination,
showing that overlap compatibility contributes to the transferable signature.
Removing the graph block \(k_{\mathrm{graph}}\) mainly affects transition-type
prediction, indicating that typed constellation structure helps separate
deformation from extension. The strong ``no constraints'' result indicates that
the constraint block \(k_{\mathrm{con}}\) is overactive for kernel
generalization in this small benchmark. Panel~(b) compares generalization
protocols: within-family and mixed-variant generalization are nearly saturated,
whereas leave-one-family-out remains the harder analogical-transfer setting.
Exact aggregate values for these secondary diagnostics are reported in
\ref{app:kernel_probe_details}.

\subsection{Qualitative case studies}

The extension-required families show how \(\Obs_S(\K_j)\) separates local fit
from representational coherence. In each case, the base or fixed-language
candidate remains meaningful in the source regime \(D_s\), but fails to give a
coherent source--overlap--target account across \(D_s,D_o,D_t\). The selected
extension lowers obstruction not only by reducing residuals, but by improving
\(G_{\mathrm{glue}}(\K_j)\), satisfying the relevant
\(C_{\mathrm{viol}}(\K_j)\), and preserving the source limit through
\(P_{\mathrm{limit}}(\K_j)\).

\paragraph{Galilean \(\rightarrow\) Lorentz velocity composition.}
The source regime \(D_s\) contains low velocities where additive composition is
locally adequate. The target regime \(D_t\) contains high subluminal velocities
where invariant-speed and speed-bound constraints become active, and the
overlap \(D_o\) tests whether the source-fitted and target-fitted charts agree.
The fixed additive candidate has total obstruction \(3.870\). A polynomial
deformation lowers the obstruction to \(0.771\), but still leaves nonzero
gluing, constraint, and limit penalties: it improves fit without fully restoring
local-to-global coherence. The Lorentzian extension has obstruction \(0.401\).
It pays representational cost \(\lambda\Cost(\Delta_j)\), but lowers the total
score by preserving the low-speed limit, satisfying the invariant-speed
constraint, and improving overlap compatibility. The selected move is therefore
not only a change of formula; it is an extension of the velocity constellation
by invariant-speed structure.

\paragraph{Newtonian \(\rightarrow\) relativistic kinetic energy.}
Here \(D_s\) is the low-speed kinetic-energy regime, while \(D_t\) approaches
relativistic velocities. The Newtonian fixed-language candidate has obstruction
\(11.315\), reflecting large target and compatibility failures when the
low-speed chart is transported too far. The best incorrect rational extension
reduces the obstruction to \(6.894\), but does not recover the full
source--target coherence. The relativistic energy extension has obstruction
\(0.389\). It reduces the high-velocity residual terms, improves the gluing
behavior on \(D_o\), and preserves the Newtonian expression as the appropriate
low-velocity limit. This case illustrates the role of \(P_{\mathrm{limit}}\):
the extension succeeds not by discarding the source theory, but by retaining it
as a limiting chart.

\paragraph{Rayleigh--Jeans \(\rightarrow\) Planck blackbody law.}
The source data \(D_s\) sample long wavelengths where the Rayleigh--Jeans chart
is locally plausible. The target and validation regimes include shorter
wavelengths where finite-energy behavior becomes decisive. The fixed-language
candidate has obstruction \(10.334\). The best incorrect polynomial repair
reduces the obstruction to \(4.742\), but does so mainly by improving local fit
while retaining a large structural penalty. The Planck-like extension has
obstruction \(0.990\). It introduces the missing quantization-scale structure,
reduces the finite-energy constraint violation, and restores a coherent
source--overlap--target description. In terms of the obstruction components,
the selected extension wins because reductions in \(R_t\),
\(G_{\mathrm{glue}}\), and \(C_{\mathrm{viol}}\) justify the added
representational cost.

Together, these cases show why the obstruction criterion is not reducible to
flexible curve fitting. Incorrect or flexible candidates can lower some
residual terms, but the selected extension is the one that lowers the full
local-to-global obstruction \(\Obs_S(\K_j)\) while preserving the source-regime
structure encoded by \(\K_0\).

\subsection{Summary of findings}

Across the 30 transition cards, the minimum-obstruction rule
\(\widehat j(T)=\argmin_j\Obs_S(\K_j)\) behaves as a finite detector of
transport versus extension. In deformation-sufficient families, low obstruction
is achieved by candidates that remain inside the source language. In
extension-required families, fixed-language candidates retain obstruction in
target fit, overlap gluing, constraints, or limits, and the selected move is an
extension of the constellation. The primary ranking gives the strongest
evidence: it achieves \(0.900\) top-1 accuracy, \(0.950\) MRR, and \(1.000\)
transition-type accuracy. Baselines and ablations show that the gain over
residual-only ranking is not simply more accurate curve fitting; it is the
ability to organize a selected move as deformation or extension and to expose
which terms \(R_s,R_o,R_t,G_{\mathrm{glue}},C_{\mathrm{viol}},
P_{\mathrm{limit}},\Cost\) explain the decision. Stress and robustness analyses
show where the criterion is stable and where finite evidence creates boundary
cases, especially in noisy virial variants. The kernel probe confirms that
\(\PhiSig(T,\Delta_j)\) and \(\psi(G_{\K_j})\) carry structured cross-family
information, while also showing that direct obstruction ranking remains the
appropriate decision rule in this benchmark.

\section{Discussion}
\label{sec:discussion}

The results support the central interpretation of the paper: scientific
theory-shift detection can be formulated as a local-to-global problem. A
candidate representation is not assessed only by its fit to a target regime. It
is assessed by whether fitted local charts restrict coherently to overlaps,
preserve source-regime limits, satisfy the constraints that define admissible
use, and justify any added representational cost. The obstruction functional
\(\Obs_S(\K_j)\) makes these requirements explicit and turns them into a
ranking criterion.

The main empirical point is not only that the intended candidate is usually
selected. It is that the selected candidate is organized as a deformation or an
extension by explicit obstruction components. In deformation-sufficient
families, low obstruction is achieved inside the original representational
language. In extension-required families, fixed-language candidates retain
obstruction in target fit, overlap compatibility, constraints, or limits, and
low obstruction appears only after the constellation is enlarged. This is the
operational distinction needed for theory-shift detection: transport is
adequate when coherence can be restored inside the source language; extension
is required when coherence requires a new representational resource.

The baseline and ablation results clarify why the obstruction criterion is not
just residual fitting. Target residual alone can select plausible candidates in
many cards, but it does not preserve the deformation-versus-extension structure
as reliably as the full obstruction score. The local-to-global terms
\(G_{\mathrm{glue}}\), \(C_{\mathrm{viol}}\), \(P_{\mathrm{limit}}\), and
\(\Cost(\Delta_j)\) make the decision interpretable: they show whether a
candidate wins because it agrees on overlaps, satisfies structural constraints,
preserves the source limit, or pays a justified cost for added language. The
point of obstruction is therefore diagnostic organization, not merely a higher
aggregate accuracy number.

The finite sheaf-theoretic language provides the organizing structure for this
diagnosis. Contexts are scientific regimes; local sections are fitted
representational constellations; restriction evaluates a chart on an overlap;
gluing measures compatibility between locally fitted descriptions; obstruction
measures failure of coherent transport. This use of sheaf language is finite
and computational, not a claim to full topos semantics. Its role is to make
local-to-global coherence testable in a controlled setting, in line with
applied sheaf-theoretic work where consistency, distributed measurements, and
global compatibility are treated as computable objects
\citep{maclane_moerdijk,johnstone,robinson2017,hansen_ghrist2019,curry2014}.

The deformation and extension cases illustrate two complementary modes of
theory shift. The deformation cases show that moving outside a source regime
does not automatically require a new language: small-angle dynamics, ideal-gas
behavior, and Ohmic response can be carried into neighboring regimes by bounded
changes that preserve the source constellation. The extension cases show the
opposite pattern. Galilean velocity addition, Newtonian kinetic energy, and
Rayleigh--Jeans radiation remain meaningful in their source regimes, but fixed
language candidates fail to give coherent source--overlap--target
descriptions. The successful candidates introduce invariant-speed structure,
relativistic energy, or a quantization scale. These are not merely better curve
fits; they change what counts as an admissible constellation of observables,
constraints, limits, and transformations, matching the role of representational
reorganization emphasized in accounts of conceptual change
\citep{kuhn,nersessian2008creating,thagard2012}.

The constellation kernel gives a secondary test of whether obstruction
signatures define a reusable representational geometry. It does not replace the
selection rule \(\Obs_S\). Instead, it asks whether
\(\PhiSig(T,\Delta_j)\) and \(\psi(G_{\K_j})\) carry similarity information
across transition families. The kernel results suggest that gluing and graph
features are transition-relevant, while the constraint block needs better
calibration for cross-family generalization. Within-family and mixed-variant
generalization are easier than leave-family-out transfer, which is the more
demanding analogical setting.

The stress and robustness analyses identify where the finite diagnostic is
stable and where it reaches boundary cases. Adding incorrect, randomized, and
matched-cost candidates does not collapse the ranking, which argues against the
simple explanation that the intended candidate wins only by having more
expressive capacity. The failures that do occur are localized, especially in
noisy virial variants where a partial low-cost correction can appear more
coherent than the intended quadratic correction. Noise also affects the ranking
more strongly than moderate subsampling, indicating that the obstruction signal
depends more on the integrity of local evidence than on having every record
available. These boundary cases are informative because they show which parts
of the diagnostic depend on data quality, cost calibration, and finite sampling.

The paper therefore sits between computational scientific discovery and
cognitive accounts of conceptual change. Computational discovery systems often
search for compact laws, symbolic expressions, or governing equations from data
\citep{langley1987scientific,schmidt2009distilling,brunton2016sindy,udrescu2020ai_feynman}.
Cognitive and philosophical accounts emphasize model-based reasoning,
conceptual restructuring, and new representational resources
\citep{nersessian2008creating,thagard2012,morgan_morrison1999}. The present
framework connects these perspectives through a finite criterion: the object of
evaluation is not only a formula, but a representational constellation; and the
failure signal is not only prediction error, but obstruction to coherent
transport.

The broader implication is that representational strain should not be identified
with a single anomalous datum. It is a pattern: local fits remain possible, but
overlap agreement, source-limit preservation, constraint satisfaction, and
representational cost cannot be jointly maintained inside the original
constellation. In this sense, extension is triggered when local adequacy no
longer supports global coherence.

\subsection{Scope, limitations, and future work}
\label{sec:limitations}

The experiment establishes a finite mechanism for theory-shift detection:
obstruction-based selection between deformation and extension. The transition
families are curated rather than historical, because the goal is to isolate the
diagnostic operation under controlled conditions. Candidates are supplied to the
ranking procedure rather than generated autonomously. Thus the benchmark is not
a reconstruction of scientific history and not a general machine-learning
benchmark. It is a controlled cognitive-systems test of whether structured
local-to-global evidence allows an artificial scientific agent to distinguish
adaptation inside a representational language from extension of that language.

The mathematical scope is also finite. Contexts, restrictions, overlap
comparisons, gluing residuals, constraints, limits, and costs are explicitly
represented and measured. The paper uses sheaf-theoretic structure as a finite
local-to-global formalism; it does not implement full topos semantics. The main
empirical boundary is the secondary kernel: direct obstruction ranking is the
stronger criterion, while the kernel remains a probe of the representation
space induced by obstruction signatures. Robustness tests also identify a data
boundary: the diagnosis is more sensitive to observation noise than to moderate
subsampling.

The next step is to expand transition cards beyond hand-designed
physics-inspired cases. A larger open transition-card database could include
curated cases across domains, explicit source, overlap, target, and validation
regimes, candidate representational moves, constraints, limits, and train/test
splits for evaluating AI agents. LLM-assisted card synthesis may be useful for
proposing source/target theory pairs, candidate deformations, plausible
incorrect extensions, constraints, and validation regimes, but such proposals
would require symbolic validation, consistency checks, and human curation before
entering a benchmark. A larger atlas of transition cards could then support
training and evaluation of AI systems for representational transport,
extension, and theory-shift detection. Richer categorical or topos-theoretic
treatments may later describe theory transport through diagrams of contexts,
common refinements, or pullback-like reconciliation structures; the present
paper supplies the finite obstruction primitive needed before that broader
program can be made operational.

\section{Conclusion}
\label{sec:conclusion}

This paper formulated scientific theory shift as a finite local-to-global
diagnostic problem. A scientific model was represented as a constellation of
observables, law schemas, constraints, measurement roles, limiting regimes, and
admissible transformations. Transport preserves this constellation by
deformation across contexts; extension enlarges it when the original language no
longer supports coherent gluing.

The obstruction functional makes this distinction computable. Source, overlap,
target, and validation regimes define the finite contextual structure. Candidate
constellations are fitted locally, restricted to overlaps, and evaluated by
residual fit, gluing disagreement, constraint violation, limit failure, and
representational cost. The selected move is therefore not simply the candidate
that best fits a target dataset, but the candidate that best restores
local-to-global coherence.

The benchmark results support this formulation as a controlled proof of
concept. Direct obstruction ranking identifies the intended deformation or
extension in most transition cards and separates deformation-sufficient from
extension-required cases in this benchmark. Baselines, ablations, stress tests,
and robustness analyses show that the diagnosis is not reducible to target
residual, raw expressive capacity, or arbitrary cost. The constellation kernel
adds a secondary representational probe, showing that obstruction signatures and
typed graph features contain transition-relevant structure, while confirming
that direct obstruction remains the main decision criterion.

The contribution is a finite computational primitive for a central cognitive
operation in scientific modeling: deciding when a representation still
transports and when obstruction motivates extension. In this view,
discovery-like revision begins where local adequacy no longer glues into global
coherence.

\bibliographystyle{model5}
\bibliography{refs}

\newpage

\appendix

\section{Auxiliary Evaluation Protocols}
\label{app:auxiliary_protocols}

The main text defines the primary procedures: transition-card construction and
finite obstruction ranking. This appendix records the auxiliary protocols used
for the secondary constellation-kernel probe, stress expansion, and robustness
sweeps. These procedures do not define additional discovery mechanisms; they
test transfer, stability, and boundary cases around the direct obstruction
ranking rule.

\begin{algorithm}[H]
\caption{Leave-family-out constellation-kernel evaluation}
\label{alg:kernel_eval}
\KwInput{Candidate signatures \(\{\PhiSig(T,\Delta_j)\}\) grouped by transition family}
\KwOutput{Kernel ranking metrics}
Construct the additive constellation kernel \(k(a,b)\) from Eq.~\eqref{eq:kernel}\;
\ForEach{held-out family \(f\in\mathcal F\)}{
  Let \(\mathcal A_{\mathrm{train}}\) contain candidate rows
  \(a=(T,\Delta_j)\) from \(\mathcal F\setminus\{f\}\)\;
  Let \(\mathcal A_{\mathrm{test}}\) contain candidate rows from \(f\)\;
  Fit the kernel scoring model on \(\mathcal A_{\mathrm{train}}\)\;
  Score and rank candidates in each held-out transition card\;
  Record top-1 accuracy, reciprocal rank, and transition-type correctness\;
}
\Return{family-level and aggregate kernel metrics}
\end{algorithm}

\begin{algorithm}[H]
\caption{Stress expansion with incorrect and matched-cost alternatives}
\label{alg:stress_expansion}
\KwInput{Transition-card collection \(\mathcal T\)}
\KwOutput{Stress-test margins and boundary cases}
\ForEach{transition card \(T=(\K_0,D_s,D_o,D_t,D_v,\{\Delta_j\}_{j=1}^m)\in\mathcal T\)}{
  Expand the candidate set with controlled incorrect formulas, randomized
  perturbations, and matched-cost incorrect extensions\;
  For each candidate move \(\Delta_j\), form \(\K_j=\Delta_j(\K_0)\) and
  compute \(\Obs_S(\K_j)\)\;
  Compute the stress margin
  \[
      M(T)
      =
      \Obs_S(\K_{\mathrm{best\ incorrect}})
      -
      \Obs_S(\K_{\mathrm{ref}}),
  \]
  where \(\K_{\mathrm{ref}}\) is the benchmark-correct candidate and
  \(\K_{\mathrm{best\ incorrect}}\) is the lowest-obstruction non-reference
  or matched-cost alternative\;
  Mark \(T\) as a boundary case if \(M(T)<0\)\;
}
\Return{stress margins, selected candidates, and boundary cases}
\end{algorithm}

\begin{algorithm}[H]
\caption{Noise and observation-record robustness sweep}
\label{alg:robustness_sweep}
\KwInput{Transition-card collection \(\mathcal T\); noise levels \(\mathcal E\);
record fractions \(\mathcal Q\)}
\KwOutput{Robustness metrics over \(\mathcal E\times\mathcal Q\)}
\ForEach{\(\eta\in\mathcal E\)}{
  \ForEach{\(q\in\mathcal Q\)}{
    Perturb the observation values in \(D_s,D_o,D_t,D_v\) at noise level \(\eta\)\;
    Retain a fraction \(q\) of observation records in each context\;
    Recompute \(\widehat j(T)=\argmin_j\Obs_S(\K_j)\) for each perturbed card\;
    Record top-1 accuracy, mean reciprocal rank, transition-type accuracy,
    and margin statistics\;
  }
}
\Return{robustness summaries over noise levels and retained-record fractions}
\end{algorithm}

\FloatBarrier

\section{Transition-Card Anatomy}
\label{app:transition_card_anatomy}

The benchmark has three nested levels: transition family, transition card, and
observation record. A transition family specifies a scientific-transition
archetype and the deformation-versus-extension question. A transition card is
one concrete instance of that family, with a source constellation, source,
overlap, target, and validation contexts, candidate moves, constraints, limits,
and an evaluation label. An observation record is one local context-indexed
datum inside one regime of the card. Obstruction is computed over the whole
transition card for each candidate move, not over a single observation record.

\begin{table}[htbp]
\centering
\small
\caption{Nested levels in the transition-card benchmark.}
\label{tab:card_hierarchy}
\begin{tabular}{@{}p{0.20\linewidth}p{0.28\linewidth}p{0.44\linewidth}@{}}
\toprule
Level & Example & Role in the benchmark \\
\midrule
Transition family
& Galilean velocity composition \(\rightarrow\) Lorentzian velocity composition
& Scientific-transition archetype specifying the source theory, target regime,
and deformation-versus-extension question. \\
Transition card
& One Galilean-to-Lorentz velocity card
& Concrete benchmark instance with \(\K_0\), \(D_s,D_o,D_t,D_v\), candidate
moves \(\{\Delta_j\}\), constraints, limits, and an evaluation label. \\
Observation record
& One sampled pair \((u,v)\) with observed composed velocity \(w\) in one context
& Local datum used to fit charts, compute residuals, test constraints, and
evaluate gluing. \\
\bottomrule
\end{tabular}
\end{table}

For example, in the Galilean-to-Lorentz family, a transition card asks whether
Galilean velocity addition can be transported into a higher-velocity regime by
bounded deformation or whether Lorentzian structure must be added. The source
data \(D_s\) contain low-velocity observations, the overlap data \(D_o\) contain
intermediate velocities, the target data \(D_t\) contain high subluminal
velocities, and \(D_v\) is held out for validation. Candidate moves include the
unchanged Galilean law, deformations inside the original language, controlled
incorrect extensions, and the Lorentzian extension. The obstruction
\(\Obs_S(\K_j)\) is computed for each candidate constellation
\(\K_j=\Delta_j(\K_0)\).

\FloatBarrier

\section{Secondary Kernel Details}
\label{app:kernel_probe_details}

The constellation kernel is a secondary diagnostic. The main text summarizes
the kernel results in Figure~\ref{fig:kernel_probe_summary}; the tables below
report the exact aggregate values behind the compact kernel diagnostics.

Table~\ref{tab:kernel_ablation} reports ablations of the kernel blocks in
Eq.~\eqref{eq:kernel}. These values are used only to interpret the secondary
kernel probe and do not replace direct obstruction ranking by \(\Obs_S\).

\begin{table}[htbp]
\centering
\small
\caption{Kernel ablation results. The ablations test which signature blocks contribute to the secondary constellation-kernel probe.}
\label{tab:kernel_ablation}
\begin{tabular}{lcccp{0.38\linewidth}}
\toprule
Ablation & Top-1 & MRR & Type & Interpretation \\
\midrule
Full kernel & 0.600 & 0.783 & 0.800 & Baseline constellation-kernel probe. \\
No gluing & 0.400 & 0.689 & 0.667 & Gluing contributes to ranking and transition-type separation. \\
No constraints & 0.867 & 0.917 & 1.000 & Constraint block is overactive for kernel generalization in this benchmark. \\
No limits & 0.600 & 0.783 & 0.800 & Limit block is neutral in this small kernel probe. \\
No graph & 0.600 & 0.767 & 0.667 & Graph features help transition-type discrimination. \\
\bottomrule
\end{tabular}
\end{table}

Table~\ref{tab:large_kernel_suite} reports a controlled variant suite for the
secondary kernel. The suite distinguishes within-family generalization,
mixed-variant generalization, and leave-family-out transfer. The last setting
is the hardest because the kernel must compare candidate signatures across
different scientific-transition archetypes.

\begin{table}[htbp]
\centering
\small
\caption{Controlled variant constellation-kernel suite. The suite probes representation quality of the secondary kernel and does not replace the primary obstruction-ranking experiment.}
\label{tab:large_kernel_suite}
\begin{tabular}{llcccc}
\toprule
Protocol & Task & Top-1 & MRR & Type & Retrieval / pairwise \\
\midrule
Leave-one-family-out & Kernel ranking & 0.673 & 0.812 & 0.847 & retrieval 0.813 \\
Leave-one-family-out & Pairwise ranking & 0.907 & 0.940 & 1.000 & preference 0.978 \\
Mixed leave-variant-out & Kernel ranking & 1.000 & 1.000 & 1.000 & retrieval 1.000 \\
Mixed leave-variant-out & Pairwise ranking & 1.000 & 1.000 & 1.000 & preference 1.000 \\
Within-family held-out variants & Kernel ranking & 1.000 & 1.000 & 1.000 & retrieval 1.000 \\
Within-family held-out variants & Pairwise ranking & 1.000 & 1.000 & 1.000 & preference 1.000 \\
\bottomrule
\end{tabular}
\end{table}

\FloatBarrier

\section{Reproducibility Note}
\label{app:reproducibility}

The accompanying implementation records candidate-level obstruction components,
stress-test margins, sensitivity sweeps, robustness summaries, and kernel
diagnostics as structured tables. The manuscript reports the corresponding
figures, aggregate metrics, and conceptual summaries; implementation-level
filenames and run artifacts remain with the accompanying code rather than the
scientific narrative.

\end{document}